\documentclass[lettersize,journal]{IEEEtran}
\usepackage{fancyhdr, epsfig, epsf, amsthm, amsmath, amssymb, amsfonts, subfigure, color}
\usepackage{threeparttable}
\usepackage[noadjust]{cite}
\usepackage{dsfont}
\usepackage{enumerate}
\usepackage{comment}
\usepackage{soul}
\usepackage{kotex}
\usepackage{algorithm}
\usepackage{algpseudocode}
\usepackage[thinc]{esdiff}
\usepackage{multirow}
\usepackage{makecell}

\makeatletter
\newcommand{\algmargin}{\the\ALG@thistlm}
\makeatother
\algnewcommand{\parState}[1]{\State%
    \parbox[t]{\dimexpr\linewidth-\algmargin}{\hangindent=\algorithmicindent \hangafter=1 #1\strut}}

\usepackage{cite}
\usepackage{graphicx}
\usepackage{textcomp}
\usepackage{xcolor}
\usepackage[scr = zapfc]{mathalfa}

\includecomment{comment}	

\newcolumntype{C}[1]{>{\centering\arraybackslash}p{#1}}

\def\BibTeX{{\rm B\kern-.05em{\sc i\kern-.025em b}\kern-.08em
    T\kern-.1667em\lower.7ex\hbox{E}\kern-.125emX}}

\usepackage{mathtools}
\usepackage[noabbrev]{cleveref}

\ExplSyntaxOn

\NewDocumentCommand \CianFormat { m }
  {
    \regex_match:nnTF { [A-Za-z] } { #1 }
      { (\textbf{#1}) }
      { (#1) }
  }

\ExplSyntaxOff

\newtagform{Cian}[\CianFormat]{}{}
\makeatletter
\renewcommand{\eqref}[1]{\textup{(\ignorespaces\ref{#1}\unskip\@@italiccorr)}}
\makeatother

\IEEEoverridecommandlockouts

\begin{document}
\usetagform{Cian}
\newtheorem{theorem}{Theorem}
\newtheorem{acknowledgement}[theorem]{Acknowledgement}
\newtheorem{axiom}[theorem]{Axiom}
\newtheorem{case}[theorem]{Case}
\newtheorem{claim}[theorem]{Claim}
\newtheorem{conclusion}[theorem]{Conclusion}
\newtheorem{condition}[theorem]{Condition}
\newtheorem{conjecture}[theorem]{Conjecture}
\newtheorem{criterion}[theorem]{Criterion}
\newtheorem{definition}{Definition}
\newtheorem{exercise}[theorem]{Exercise}
\newtheorem{lemma}{Lemma}
\newtheorem{corollary}{Corollary}
\newtheorem{notation}[theorem]{Notation}
\newtheorem{problem}[theorem]{Problem}
\newtheorem{proposition}{Proposition}
\newtheorem{remark}{Remark}
\newtheorem{solution}[theorem]{Solution}
\newtheorem{summary}[theorem]{Summary}
\newtheorem{assumption}{Assumption}
\newtheorem{example}{\bf Example}
\newtheorem{probform}{\bf Problem}

\def\qed{$\Box$}
\def\QED{\mbox{\phantom{m}}\nolinebreak\hfill$\,\Box$}
\def\proof{\noindent{\emph{Proof:} }}
\def\poof{\noindent{\emph{Sketch of Proof:} }}
\def
\endproof{\hspace*{\fill}~\qed
\par
\endtrivlist\unskip}
\def\endproof{\hspace*{\fill}~\qed\par\endtrivlist\vskip3pt}

\def\E{\mathsf{E}}
\def\eps{\varepsilon}
\def\Lsp{{\boldsymbol L}}
\def\Bsp{{\boldsymbol B}}
\def\lsp{{\boldsymbol\ell}}
\def\Ltsp{{\Lsp^2}}
\def\Lpsp{{\Lsp^p}}
\def\Linsp{{\Lsp^{\infty}}}
\def\LtR{{\Lsp^2(\Rst)}}
\def\ltZ{{\lsp^2(\Zst)}}
\def\ltsp{{\lsp^2}}
\def\ltZt{{\lsp^2(\Zst^{2})}}
\def\ninN{{n{\in}\Nst}}
\def\oh{{\frac{1}{2}}}
\def\grass{{\cal G}}
\def\ord{{\cal O}}
\def\dist{{d_G}}
\def\conj#1{{\overline#1}}
\def\ntoinf{{n \rightarrow \infty }}
\def\toinf{{\rightarrow \infty }}
\def\tozero{{\rightarrow 0 }}
\def\trace{{\operatorname{trace}}}
\def\ord{{\cal O}}
\def\UU{{\cal U}}
\def\rank{{\operatorname{rank}}}
\def\acos{{\operatorname{acos}}}

\def\SINR{\mathsf{SINR}}
\def\SNR{\mathsf{SNR}}
\def\SIR{\mathsf{SIR}}
\def\tSIR{\widetilde{\mathsf{SIR}}}
\def\Ei{\mathsf{Ei}}
\def\l{\left}
\def\r{\right}
\def\({\left(}
\def\){\right)}
\def\lb{\left\{}
\def\rb{\right\}}

\setcounter{page}{1}

\newcommand{\eref}[1]{(\ref{#1})}
\newcommand{\fig}[1]{Fig.\ \ref{#1}}

\def\bydef{:=}
\def\ba{{\mathbf{a}}}
\def\bb{{\mathbf{b}}}
\def\bc{{\mathbf{c}}}
\def\bd{{\mathbf{d}}}
\def\bee{{\mathbf{e}}}
\def\bff{{\mathbf{f}}}
\def\bg{{\mathbf{g}}}
\def\bh{{\mathbf{h}}}
\def\bi{{\mathbf{i}}}
\def\bj{{\mathbf{j}}}
\def\bk{{\mathbf{k}}}
\def\bl{{\mathbf{l}}}
\def\bm{{\mathbf{m}}}
\def\bn{{\mathbf{n}}}
\def\bo{{\mathbf{o}}}
\def\bp{{\mathbf{p}}}
\def\bq{{\mathbf{q}}}
\def\br{{\mathbf{r}}}
\def\bs{{\mathbf{s}}}
\def\bt{{\mathbf{t}}}
\def\bu{{\mathbf{u}}}
\def\bv{{\mathbf{v}}}
\def\bw{{\mathbf{w}}}
\def\bx{{\mathbf{x}}}
\def\by{{\mathbf{y}}}
\def\bz{{\mathbf{z}}}
\def\b0{{\mathbf{0}}}

\def\bA{{\mathbf{A}}}
\def\bB{{\mathbf{B}}}
\def\bC{{\mathbf{C}}}
\def\bD{{\mathbf{D}}}
\def\bE{{\mathbf{E}}}
\def\bF{{\mathbf{F}}}
\def\bG{{\mathbf{G}}}
\def\bH{{\mathbf{H}}}
\def\bI{{\mathbf{I}}}
\def\bJ{{\mathbf{J}}}
\def\bK{{\mathbf{K}}}
\def\bL{{\mathbf{L}}}
\def\bM{{\mathbf{M}}}
\def\bN{{\mathbf{N}}}
\def\bO{{\mathbf{O}}}
\def\bP{{\mathbf{P}}}
\def\bQ{{\mathbf{Q}}}
\def\bR{{\mathbf{R}}}
\def\bS{{\mathbf{S}}}
\def\bT{{\mathbf{T}}}
\def\bU{{\mathbf{U}}}
\def\bV{{\mathbf{V}}}
\def\bW{{\mathbf{W}}}
\def\bX{{\mathbf{X}}}
\def\bY{{\mathbf{Y}}}
\def\bZ{{\mathbf{Z}}}

\def\mA{{\mathbb{A}}}
\def\mB{{\mathbb{B}}}
\def\mC{{\mathbb{C}}}
\def\mD{{\mathbb{D}}}
\def\mE{{\mathbb{E}}}
\def\mF{{\mathbb{F}}}
\def\mG{{\mathbb{G}}}
\def\mH{{\mathbb{H}}}
\def\mI{{\mathbb{I}}}
\def\mJ{{\mathbb{J}}}
\def\mK{{\mathbb{K}}}
\def\mL{{\mathbb{L}}}
\def\mM{{\mathbb{M}}}
\def\mN{{\mathbb{N}}}
\def\mO{{\mathbb{O}}}
\def\mP{{\mathbb{P}}}
\def\mQ{{\mathbb{Q}}}
\def\mR{{\mathbb{R}}}
\def\mS{{\mathbb{S}}}
\def\mT{{\mathbb{T}}}
\def\mU{{\mathbb{U}}}
\def\mV{{\mathbb{V}}}
\def\mW{{\mathbb{W}}}
\def\mX{{\mathbb{X}}}
\def\mY{{\mathbb{Y}}}
\def\mZ{{\mathbb{Z}}}

\def\cA{\mathcal{A}}
\def\cB{\mathcal{B}}
\def\cC{\mathcal{C}}
\def\cD{\mathcal{D}}
\def\cE{\mathcal{E}}
\def\cF{\mathcal{F}}
\def\cG{\mathcal{G}}
\def\cH{\mathcal{H}}
\def\cI{\mathcal{I}}
\def\cJ{\mathcal{J}}
\def\cK{\mathcal{K}}
\def\cL{\mathcal{L}}
\def\cM{\mathcal{M}}
\def\cN{\mathcal{N}}
\def\cO{\mathcal{O}}
\def\cP{\mathcal{P}}
\def\cQ{\mathcal{Q}}
\def\cR{\mathcal{R}}
\def\cS{\mathcal{S}}
\def\cT{\mathcal{T}}
\def\cU{\mathcal{U}}
\def\cV{\mathcal{V}}
\def\cW{\mathcal{W}}
\def\cX{\mathcal{X}}
\def\cY{\mathcal{Y}}
\def\cZ{\mathcal{Z}}
\def\cd{\mathcal{d}}
\def\Mt{M_{t}}
\def\Mr{M_{r}}
\def\O{\Omega_{M_{t}}}
\newcommand{\figref}[1]{{Fig.}~\ref{#1}}
\newcommand{\tabref}[1]{{Table}~\ref{#1}}

\newcommand{\var}{\mathsf{var}}
\newcommand{\fb}{\tx{fb}}
\newcommand{\nf}{\tx{nf}}
\newcommand{\BC}{\tx{(bc)}}
\newcommand{\MAC}{\tx{(mac)}}
\newcommand{\Pout}{p_{\mathsf{out}}}
\newcommand{\nnn}{\nn\\}
\newcommand{\FB}{\tx{FB}}
\newcommand{\TX}{\tx{TX}}
\newcommand{\RX}{\tx{RX}}
\renewcommand{\mod}{\tx{mod}}
\newcommand{\m}[1]{\mathbf{#1}}
\newcommand{\td}[1]{\tilde{#1}}
\newcommand{\sbf}[1]{\scriptsize{\textbf{#1}}}
\newcommand{\stxt}[1]{\scriptsize{\textrm{#1}}}
\newcommand{\suml}[2]{\sum\limits_{#1}^{#2}}
\newcommand{\sumlk}{\sum\limits_{k=0}^{K-1}}
\newcommand{\eqhsp}{\hspace{10 pt}}
\newcommand{\tx}[1]{\texttt{#1}}
\newcommand{\Hz}{\ \tx{Hz}}
\newcommand{\sinc}{\tx{sinc}}
\newcommand{\tr}{\mathrm{tr}}
\newcommand{\diag}{\mathrm{diag}}
\newcommand{\MAI}{\tx{MAI}}
\newcommand{\ISI}{\tx{ISI}}
\newcommand{\IBI}{\tx{IBI}}
\newcommand{\CN}{\tx{CN}}
\newcommand{\CP}{\tx{CP}}
\newcommand{\ZP}{\tx{ZP}}
\newcommand{\ZF}{\tx{ZF}}
\newcommand{\SP}{\tx{SP}}
\newcommand{\MMSE}{\tx{MMSE}}
\newcommand{\MINF}{\tx{MINF}}
\newcommand{\RC}{\tx{MP}}
\newcommand{\MBER}{\tx{MBER}}
\newcommand{\MSNR}{\tx{MSNR}}
\newcommand{\MCAP}{\tx{MCAP}}
\newcommand{\vol}{\tx{vol}}
\newcommand{\ah}{\hat{g}}
\newcommand{\tg}{\tilde{g}}
\newcommand{\teta}{\tilde{\eta}}
\newcommand{\heta}{\hat{\eta}}
\newcommand{\uh}{\m{\hat{s}}}
\newcommand{\eh}{\m{\hat{\eta}}}
\newcommand{\hv}{\m{h}}
\newcommand{\hh}{\m{\hat{h}}}
\newcommand{\Po}{P_{\mathrm{out}}}
\newcommand{\Poh}{\hat{P}_{\mathrm{out}}}
\newcommand{\Ph}{\hat{\gamma}}
\newcommand{\mat}[1]{\begin{matrix}#1\end{matrix}}
\newcommand{\ud}{^{\dagger}}
\newcommand{\C}{\mathcal{C}}
\newcommand{\nn}{\nonumber}
\newcommand{\nInf}{U\rightarrow \infty}

\title{Energy-Efficient Edge Learning via \\ Joint Data Deepening-and-Prefetching}
\author{Sujin Kook, Won-Yong Shin, Seong-Lyun Kim, and Seung-Woo Ko, \emph{Senior Member}, \emph{IEEE}
\thanks{This work was supported by Institute of Information \& communications Technology Planning \& Evaluation (IITP) grant funded by the Korea government (MSIT) (No.2021-0-00270, Development of 5G MEC framework to improve food factory productivity, automate and optimize flexible packaging and No.2021-0-00347, 6G Post-MAC (POsitioning- \& Spectrum-aware intelligenT MAC for Computing \& Communication Convergence)) and by the National Research Foundation of Korea (NRF) grant funded by the Korea government (MSIT) under Grant RS-2023-00220762. \emph{(Corresponding author : Seung-Woo Ko.)}

S. Kook and S.-L. Kim are with the Department of Electrical and Electronics Engineering, Yonsei University, Seoul 03722, South Korea (email: \{sjkook, slkim\}@ramo.yonsei.ac.kr). 
W.-Y. Shin is with the School of Mathematics and Computing (Computational Science and Engineering), Yonsei University, Seoul 03722, South Korea, and also with the Graduate School of Artificial Intelligence, Pohang University of Science and Technology (POSTECH), Pohang 37673, South Korea (email: wy.shin@yonsei.ac.kr).
S.-W. Ko is with the Department of Smart Mobility Engineering, Inha University, Incheon 21999, South Korea (email: swko@inha.ac.kr). This paper was presented in part at IEEE ICC 2023 \cite{kook2022joint}.}
}

\maketitle

\begin{abstract}
The vision of pervasive \emph{artificial intelligence} (AI) services can be realized by training an AI model on time using real-time data collected by \emph{internet of things} (IoT) devices. To this end, IoT devices require offloading their data to an edge server in proximity. However, transmitting high-dimensional and voluminous data from energy-constrained IoT devices poses a significant challenge. To address this limitation, we propose a novel offloading architecture, called \emph{joint data deepening-and-prefetching} (JD2P), which is feature-by-feature offloading comprising two key techniques. The first one is \emph{data deepening}, where each data sample's features are sequentially offloaded in the order of importance determined by the data embedding technique such as \emph{principle component analysis} (PCA). Offloading is terminated once the already transmitted features are sufficient for accurate data classification, resulting in a reduction in the amount of transmitted data. The criteria to offload data are derived for binary and multi-class classifiers, which are designed based on \emph{support vector machine} (SVM) and \emph{deep neural network} (DNN), respectively. The second one is \emph{data prefetching}, where some features potentially required in the future are offloaded in advance, thus achieving high efficiency via precise prediction and parameter optimization. We evaluate the effectiveness of JD2P through experiments using the MNIST dataset, and the results demonstrate its significant reduction in expected energy consumption compared to several benchmarks without degrading learning accuracy.
\end{abstract}

\begin{IEEEkeywords}
Edge learning, energy efficiency, feature importance, principe component analysis, data deepening, data prefetching, support vector machine, deep neural network.
\end{IEEEkeywords}


\section{Introduction}

The vision of smart cities is to automate a wide range of services such as autonomous driving, remote control, and health care using a large volume of real-time data collected by \emph{internet-of-things} (IoT) devices pervasive in our society. With recent advancements in artificial intelligence (AI) techniques, these services can be designed in a user-customized manner by training an AI model using a target user’s data before going out of date. A new paradigm of \emph{edge learning} has emerged as a viable solution such that an AI model can be quickly trained at the edge server collocated with a wireless access point instead of a central cloud \cite{deng2020edge}. One key prerequisite of edge learning is that the data required for training should be offloaded to the edge server on time over a wireless channel, which can be a heavy burden to energy-constrained IoT devices \cite{park2021communication}. To cope with this issue, we propose a novel technique, called \emph{joint data deepening-and-prefetching} (JD2P). By leveraging a data embedding technique, each data sample can be represented by several features. A few of them likely to be critical for the concerned AI model are proactively offloaded via feature-level prediction, enabling an IoT device to reduce the amount of offloading data and enlarge the offloading duration. The resultant transmit energy efficiency can be significantly improved.

\subsection{Prior Work}\label{subsec:priorwork}

In the literature, edge learning can be categorized into two directions depending on who is in charge of training. The first one is centralized edge learning, where an edge server trains an AI model using the offloaded data samples from IoT devices \cite{mao2017survey}. On the other hand, the second one is federated edge learning, where the central server helps exchange the local AI model trained by each IoT device \cite{mcmahan2017communication}. Both directions have pros and cons, and an appropriate direction should be chosen depending on the specific application requirements and resource constraints. For example, centralized edge learning is more suitable to our concerned scenarios with an IoT device whose computation capability is insufficient to train an AI model. On the other hand, communication overhead remains a significant bottleneck for both directions to deliver data samples or AI model parameters, calling for designing an energy-efficient offloading technique as a main research~thrust \cite{jiang2020energy}. 

It is apparent that offloading fewer data facilitates to design an energy-efficient edge learning approach by saving an IoT device's transmit energy. To this end, one straightforward approach is to compress raw data as much as possible without losing the meaning embedded therein. From the perspective of data science, data embedding, \textit{a.k.a.} dimensionality reduction, is one effective data compression technique to represent a high-dimensional data sample to a few features that can be translated to a small number of bits (see, e.g., \cite{du2018fast} and \cite{azar2019energy}). Besides, quantization techniques extensively studied in the design of limited feedback can be applied to map raw data distributed on a continuous domain into a discrete domain expressable with finite bits, such as \cite{du2020high} and \cite{oh2021quantized}. The techniques mentioned above focus on encoding individual data effectively without compromising the performance of AI models when all encoded data are used for training. In other words, the contribution of individual data to the learning performance is not considered, which may seem insignificant when a large amount of data is present. On the other hand, edge learning should operate with a limited number of data samples supplied through a wireless link. It is thus vital to evaluate the effect of each data sample on the learning performance, referred to as \emph{data importance}~\cite{zhu2020toward}. 

A few recent studies have incorporated data importance into the design of communication protocols for edge learning. In \cite{liu2020data}, for example, the importance of each data sample is defined as the uncertainty of the corresponding learning result, which is quantified by the distance to the decision boundary or entropy when \emph{support vector machine} (SVM) or \emph{deep neural network} (DNN) is used, respectively. Along with an IoT device's channel condition, the data importance metric can be utilized to control retransmission, where data samples with higher uncertainty are repeatedly transmitted for clarity, improving learning accuracy. The work is extended into designing a scheduling algorithm in \cite{liu2020wireless} such that a device with the most important data is granted to transmit its data to the edge server with the highest priority. In \cite{zhi2022data}, the importance of each data sample is defined using cosine similarity. The data sample with the highest similarity is considered representative data of the corresponding dataset at the IoT device. The edge server selects the IoT device whose representative data sample differs from the server's dataset. In \cite{he2020importance}, an IoT device quantifies each data sample’s importance based on the size of the gradient vector. The edge server uses the importance metric to control the number of data samples to participate in backward propagation, thereby reducing the local training latency in the federated edge learning system. 
In \cite{taik2021data}, the importance of an IoT device's local dataset is defined from a data diversity perspective, where the dataset comprising a large number of samples with diverse classes is considered more important. The diversity metric is incorporated into a federated edge learning scheduling problem by formulating a multi-objective optimization addressing the tradeoff between fairness and learning accuracy. In \cite{ren2020scheduling}, the local AI model's importance is determined based on its gradient. An IoT device with a more significant gradient norm, which is more likely to contribute significantly to a global AI model, can send its local AI model to the edge server unless its channel condition is poor.

The above studies on data importance-aware edge learning attempted to reduce the amount of transmitting data, yet consuming significant energy if radio resources such as time and bandwidth are limited given. For energy efficiency, it is also essential to secure sufficient radio resources through a joint communication-and-computation design. In \cite{shi2020joint}, a joint device selection-and-bandwidth allocation is studied for accelerating federated learning, where a few IoT devices participating in the training are allocated bandwidth in inverse proportion to their channels. Each IoT device's transmission duration becomes balanced, and the edge server immediately updates a global model without waiting for a straggler. In \cite{wen2022federated}, the edge server generates multiple sub-models using the dropout method. This helps reduce communication and computation overhead due to a smaller model size compared to the original one. Learning latency is improved as the server determines the optimal dropout rate, considering the resources of the edge device. In \cite{ko2017live}, data prefetching is considered for energy-efficient edge computing, where partial data required for a subsequent task are proactively offloaded before being requested based on a task-level prediction. The resulting offloading duration is extended, thereby saving energy consumption. The energy-efficient federated edge learning is studied in \cite{yang2020energy}, where the overall energy consumption, including both communication and computation energies, is minimized under a delay constraint by jointly optimizing various parameters such as bandwidth, transmit power, and CPU cycle. In \cite{merluzzi2021wireless}, the optimal tradeoff among energy, delay, and accuracy is achieved by minimizing the energy consumption or maximizing the learning accuracy while constraining the others with proper designs of source quantization, CPU frequency, transmission rate, and so forth. 

Although the afore-mentioned studies cope with a wide range of energy-efficient edge learning issues, two technical challenges have not been addressed yet. First, raw data collected by an IoT device are high-dimensional in many scenarios, e.g., high-definition images and videos for surveillance services, thus resulting in a significant amount of energy to offload even though a limited number of data is sampled according to a data importance metric. In other words, measuring importance in smaller units than data is necessary, e.g., features obtained through data embedding. In \cite{lan2021progressive}, feature-importance-based offloading techniques are proposed, where an IoT device progressively transmits features until the expected gain on inference accuracy is no less than the transmission cost. However, to the best of our knowledge, there is no work addressing the \textit{feature importance} in the training process of edge learning.

Second, most edge learning scenarios include intermediate training processes to make subsequent decisions. As training an AI model takes longer due to its high computational complexity, the remaining time after the training period is reduced. A higher-rate communication should be made to deliver the required data within a short duration, making the energy consumption significant. It can be overcome by prefetching or caching data in advance, which is expected to be necessary in the future. A few \textit{proactive offloading} designs tackling this issue are proposed in the area of general computing scenarios (see, e.g., \cite{ko2017live}, \cite{elbamby2017proactive}, and \cite{ko2022computation}), while there has yet to be a study in edge learning.

\subsection{Contributions and Organizations}

This work proposes J2DP, which is a novel offloading design addressing the two limitations mentioned above for energy-efficient edge learning. Specifically, J2DP is based on two essential techniques, namely, \emph{data deepening} and \emph{data prefetching}. Specifically, JD2P is designed to determine each data sample's depth, defined as the number of features required to achieve a given training accuracy. With the help of data embedding such as \emph{principle component analysis} (PCA), high-dimensional raw data can be represented as a few features listed in order of importance. We can offload each data sample's top features according to its depth. Exploring each data's depth is made through a feature-by-feature investigation such that an intermediate learning result trained through features offloaded so far determines whether each data sample's next feature is necessary. It is called data deepening. As a result, the number of data samples offloading their following features decreases with the investigation depth, reducing the amount of offloading data. The classifiers trained through data deepening can be used to predict new data samples with their hierarchical architecture according to the feature. Second, several data samples that are highly likely to be necessary for the subsequent feature investigation can be proactively offloaded during the current training duration. It is called data prefetching, allowing the IoT device to extend the offloading duration. Through such two-fold gains, it is possible to minimize the overall energy consumption under the limited opportunities of depth investigations. The main contributions of this work are summarized as follows.

\begin{itemize}
    \item \textbf{Data Deepening for the Binary SVM Classifier}: Consider the case of a binary classifier. When the $k$-th features of several data samples are offloaded, the edge server trains a $k$-depth classifier using a classic SVM technique. We define each data sample's clarity metric as the distance to the hyperplane, determining whether its $(k+1)$-th feature is required for clarity. The decision boundary is derived using the Mahalanobis distance.

    \item \textbf{Data Deepening for the Multi-Class DNN Classifier}: 
    Consider the case of a multi-class classifier. Given the $k$-th offloaded features, the edge server trains a $k$-depth classifier using a DNN. Each data sample's clarity metrics are defined in two ways, namely the negative entropy and the gap of posteriors, using the output layer's posterior distribution. The decision boundary for determining which data samples require $(k+1)$-th feature is derived using the empirical error distribution.

    \item \textbf{Optimization of Data Prefetching \& Energy Efficiency Analysis}: When too many data samples' features are prefetched, the resultant energy consumption is more excessive than expected while some of them are not used in the subsequent classifier design. It is thus essential to balance between the gain of a longer transmission duration and the loss due to wasting energy. By solving the problem of minimizing the expected energy consumption, we derive the optimal number of prefetched data samples in a closed form. The expected energy efficiency of JD2P is compared with the benchmark offloading every data's features, showing that JD2P is more energy efficient than the benchmark, except for a highly long training duration.
\end{itemize}
Note that compared to our earlier work in \cite{kook2022joint}, we  include a new design of data deepening for a multi-class DNN classifier and the analysis of  J2DP's energy efficiency.

The remainder of this paper is organized as follows. The system model is presented in Section \ref{section:system model}. The JD2P architecture and metric of clarity are described in Section \ref{Section:overview}. The threshold for data deepening is designed in Section \ref{subsection:threshold_design}. The optimal number of prefetching data is derived in Section \ref{section:optimal_data_prefetching}. Considerations for extending the proposed algorithm are described in Section \ref{section:extension_to_general_cases}. Simulation results are presented in Section \ref{Section:Simulation} followed by concluding remarks in Section \ref{Section:concluding remarks}.

\section{System Model} \label{section:system model}

This section describes our system model, including the concerned scenario, data structure, and offloading model.  

\begin{figure}[t]
\centering
\centering
\includegraphics[width=8.3cm]{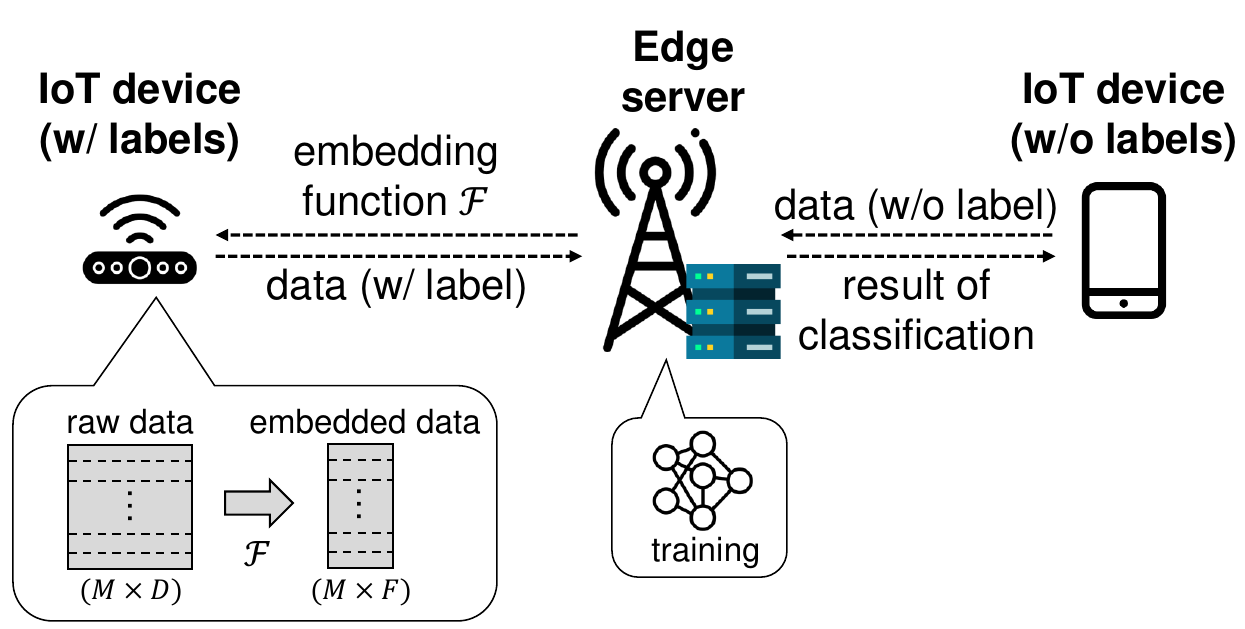}
\vspace{-5pt}
\caption{The edge learning network comprising IoT devices and an edge server collocated with a wireless access point.}
\label{Fig1-system_model}
\end{figure}

\subsection{Edge Learning Scenario}

Consider an edge learning network comprising IoT devices\footnote{One of the IoT devices is in charge of collecting data samples from its surroundings and offloading them to the edge server for training an AI model, which is the main target. On the other hand, the other without labels downloads the trained AI model and utilizes it to classify the unlabeled data.} and an edge server (see Fig. \ref{Fig1-system_model}). The objective is to train a classifier using the local data collected by the IoT device. However, the IoT device's limited computation capability makes it infeasible to train the classifier on the IoT device itself. Therefore, the local data collected by the IoT device are offloaded to the edge server whose take is to train the classifier.

\subsection{Data Embedding}\label{subsection:data_embedding}

Consider $M$ data samples measured by the sensor, denoted by $\mathbf{d}_m\in\mathbb{R}^D$, where $m$ is the index of the data sample ($m=1,2,\cdots, M$). We assume that each sample's class is known and denoted by ${c}_m \in \mathbb{C}$, where $\mathbb{C} = \{0, 1, \cdots, C-1\}$ is the set of all possible classes, and $C$ is the number of classes in $\mathbb{C}$, i.e.,  $C=|\mathbb{C}|$. Each data sample's raw dimension, denoted by $D$, is assumed to be equivalent. The dimension $D$ is generally high enough to address complex environments, yet presenting a challenge to achieving high-accuracy classification \cite{domingos2012few}. Besides, offloading a large volume of high-dimensional raw data to the edge server requires significant energy consumption. To overcome these challenges, data embedding techniques can be used to map high-dimensional raw data to a low-dimensional feature space \cite{zheng2014coupled}. Examples of these techniques include PCA \cite{abdi2010principal} and auto-encoder \cite{hinton2006reducing}. 
Specifically, given $F$ less than $D$, there exists a mapping function $\mathcal{F}:\mathbb{R}^D\rightarrow \mathbb{R}^F$ such that 
\begin{align}\label{Eq:Data_Embedding}
    \mathbf{x}_m=\mathcal{F}(\mathbf{d}_m),
\end{align}
where $\mathbf{x}_m= \left[x_{m,1}, \cdots , x_{m,F}\right]^T$ represents the embedded data with $F$ features. We assume that the embedding function $\mathcal{F}$ has been trained by the edge server using the historical dataset. The IoT device knows $\mathcal{F}$ in advance by downloading it during an off-duration. We use PCA as the primary feature embedding function $\mathcal{F}$. The main reason is not only its low computational overhead but also additional information about feature importance due to its interpretability. Specifically, PCA enables us to sort all features based on the corresponding eigenvalue’s magnitude, which is considered as the order of feature importance throughout the work. On the other hand, other representative embedding functions, e.g., autoencoder, make it difficult to quantify feature importance in a straightforward manner. It is possible to use another embedding technique as long as there is any supportable algorithm to determine such feature importance. Partial or all features of each embedded data are offloaded depending on the offloading and learning designs introduced in the sequel.

\subsection{Offloading Model}\label{subsection:offloading_model}

We only consider the IoT sensor’s energy consumption to offload data since the embedding function in \eqref{Eq:Data_Embedding} can be downloaded from the edge server\footnote{{When data distribution changes rapidly, the energy cost at the edge server could be considered for training the embedding function, which remains for future work.}}.
The entire offloading duration is slotted into $K$ rounds, each with a duration of $t_0$ seconds. The channel gain in round $k$ is denoted as $h_k>0$. We assume that channel gains remain constant within each round and are \emph{independently and identically distributed} (i.i.d.) across different rounds. Following the models in \cite{tao2019stochastic}, \cite{E2_zhang2013energy}, the transmission power required to transmit $b$ bits in round $k$, denoted as $e_k$, is modeled by a {monomial function} and given as $e_k = \lambda \frac{{(b/t)}^{\ell}}{h_k}$, where $\lambda$ is the energy coefficient, $\ell$ represents the monomial order, and $t$ is an allowable transmission duration for $b$ bits (i.e., $t\leq t_0$). The typical range for a monomial order is $2\leq \ell \leq 5$ and depends on the concerned modulation and coding schemes. The energy consumption in round $k$, given by the product of $e_k$ and $t$, is 
\begin{align} \label{EQ : E}
    \mathbf{E}(b, t;h_k) = {e_k}{t} = \lambda \frac{{b}^\ell}{h_k\left(t\right)^{\ell-1}}.
\end{align}
One observes that energy consumption is directly proportional to the transmitted data size $b$ and inversely proportional to the transmission time $t$. To achieve energy-efficient edge learning, it is necessary to decrease the amount of transmitted data and increase the transmission time.

\section{Joint Data Deepening-and-Prefetching: Overview}\label{Section:overview}
This section presents an overview of a novel JD2P architecture to realize energy-efficient edge learning. A key definition required to understand JD2P is introduced first, and two core techniques of JD2P are elaborated next.  

\subsection{Data Depth}
 
The proposed JD2P is a feature-by-feature offloading design to selectively offload a data sample's features that are likely to be more effective in training a classifier. According to the principle of PCA, it is reasonable to consider that the embedding order is equivalent to the effectiveness order; namely, data sample $m$'s $i$-th feature $x_{m,i}$ is more effective than the $j$-th feature $x_{m,j}$ on enhancing the classifier accuracy when $i<j$. Under the assumption, we introduce the following definition, which is essential to differentiate the amount of offloading data across~samples. 
    
\begin{definition}[Data Depth] \label{Defintion:DataDepth}\emph{A embedded data sample $\mathbf{x}_m$ is said to have depth $k$ when features from $1$ to $k$, say $\mathbf{x}_m^{(k)}=[x_{m,1},\cdots, x_{m,k}]^T$, are enough to correctly predict its class.}
\end{definition}

By Definition \ref{Defintion:DataDepth}, we can offload less amount of data required to train the classifier, and the resultant energy consumption can be reduced if depths of all data are known in advance. On the other hand, each data sample's depth is unknown and can be determined after the concerned classifier is trained. Eventually, it is required to perform each data's depth identification and classifier training simultaneously to cope with the above recursive relation, which is technically challenging. To this end, we propose two key techniques summarized below.  

\begin{remark}[Data Importance and Feature Importance]\emph{{
As aforementioned, there are several definitions of data importance in the literature,
ranging from the size of the gradient vector \cite{he2020importance} to information-theoretic entropy \cite{liu2020data}, most of which are measured in a unit of a data sample. On the other hand, we define feature importance, allowing us to sort features in the order of their significance to the classifier training. Along with the offloading design introduced in the sequel, it is possible to achieve better energy efficiency by offloading a few features of each data sample at the appropriate time.
}}
\end{remark}

\begin{figure}[t] 
\centering
\centering
\includegraphics[width=6.5cm]{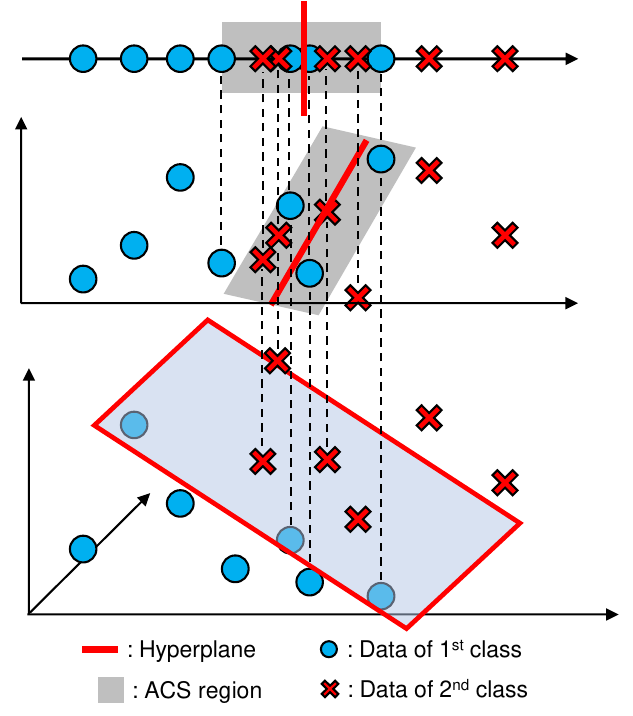}
\vspace{-5pt}
\caption{An example of data deepening from the $1$-dimensional space to the $3$-dimensional space. The data samples in the gray area are included in the ACS set. Only data samples in the ACS take into account offloading following features marked by dotted lines in the subsequent~rounds.}
\label{Fig-Data deepening}
\end{figure}

\subsection{Data Deepening}
It is a closed-loop control that determines whether a new feature should be offloaded or not based on the current version of a classifier. Specifically, define a $k$-depth classifier that is trained using features from $1$ to $k$, denoted as $\mathbf{x}_m^{(k)}$ for all $m\in \mathbb{S}^{(k)}$ where $\mathbb{S}^{(k)}$ denotes the set of candidate data samples whose depths are no less than $k$. 
We use a classic SVM and DNN for each depth classifier of binary and multi-class classifications, respectively.   
For each case, the trained classifier is used to divide the data samples in $\mathbb{S}^{(k)}$ into two groups based on the clarity of their classification, as quantified by the  metric denoted by $\beta_m^{(k)}$. It is called a \emph{metric of clarity} (MoC) throughout the paper, whose detailed form is specified in Section. \ref{subsection:metric}. A data sample with a higher MoC is said to be a \emph{clearly classified sample} (CCS), which corresponds to the depth-$k$ data sample not requiring an additional feature. On the other hand, the data sample is said to be an \emph{ambiguous classified sample} (ACS), which will be included in a new candidate set $\mathbb{S}^{(k+1)}$. 
Denote $\bar{\beta}^{(k)}$ the boundary determining between CCS and ACS specified in Section. \ref{subsection:threshold_design}. Then, $\mathbb{S}^{(k+1)}$ is given as
\begin{align}\label{eq : ACI set}
    \mathbb{S}^{(k+1)} = \left\{ m~ |~ \beta_m^{(k)} \leq \bar{\beta}^{(k)}, m\in \mathbb{S}^{(k)}\right\}.
\end{align}
The derivation of the threshold $\bar{\beta}^{(k)}$ will be elaborated in Section. \ref{subsection:threshold_design}.

\begin{algorithm}[t]
\caption{Data Deepening}
\begin{algorithmic}[1]
    \Require Embedded data sample $\mathbf{x}_m$ for all $m\in\{1,\cdots, M\}$.
    \State Setting $k = 0$, $\mathbb{S}^{(1)}=\{1,\cdots,M\} $.
    \While{$k \leq K$}
    \State $k = k + 1$.
    \parState{%
    Using $\{\mathbf{x}_m^{(k)}\}$ for $m \in \mathbb{S}^{(k)}$, train the $k$-depth classifier.}
    \State Compute the threshold $\bar{\beta}_m^{(k)}$ of clarity. 
    \For {$m \in \mathbb{S}^{(k)}$}
        \State Compute the clarity $\beta_m^{(k)}$ according to classifier.
        \If{$\beta_m^{(k)} \leq \bar{\beta}^{(k)}$}.
            \State $m \in \mathbb{S}^{(k+1)}$.
        \Else
            \State $m \notin \mathbb{S}^{(k+1)}$.
        \EndIf
    \EndFor
    \EndWhile
\end{algorithmic}
\label{algorithm:deepening}
\end{algorithm}

When the set of ACSs, say $\mathbb{S}^{(k+1)}$ of \eqref{eq : ACI set}, is determined, the edge server requests the {IoT device} to offload the next feature $x_{m,k+1}$ for all $m\in\mathbb{S}^{(k+1)}$ to train the $(k+1)$-depth classifier. Fig. \ref{Fig-Data deepening} illustrates a graphical example of data deepening from $1$-dimensional to $3$-dimensional feature spaces. ACSs are represented as points vertically connected by dotted lines, requiring an additional feature to get detailed information. Conversely, CCSs are represented as disconnected ones and do not need further features. The detailed process of data deepening is summarized in~Algorithm~\ref{algorithm:deepening}.


\subsection{Data Prefetching} 
\begin{figure}[t]
\centering
\includegraphics[width=9cm]{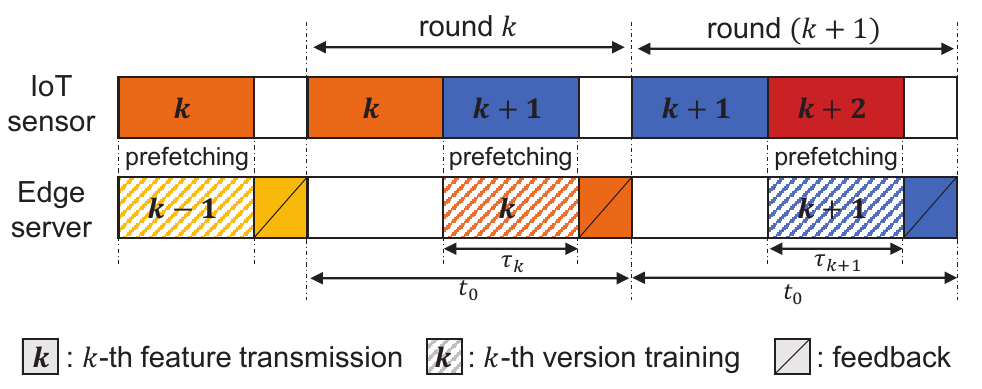}
\caption{The data prefetching architecture. Each round comprises offloading duration, training duration, and feedback duration. The $(k+1)$-th ACS set, $\mathbb{S}^{(k+1)}$, can be available when the round $(k+1)$ starts. A few data samples' $(k+1)$-th features,  $\mathbf{x}_{m,k+1}$, are prefetched from the {IoT device} to the edge server during the training duration of the $k$-th depth classifier. The remaining data samples' $(k+1)$-th features are offloaded after training if they belong to the $(k+1)$-th ACS set.}  
\label{Fig-prefetching}
\end{figure}

As shown in Fig. \ref{Fig-prefetching}, the round $k$ comprises  an offloading duration for the $k$-th features (i.e., $x_{m,k}, \forall m\in\mathbb{S}^{(k)}$), a training duration for the $k$-depth classifier, and a feedback duration for a new ACS set 
$\mathbb{S}^{(k+1)}$ in \eqref{eq : ACI set}. Without loss of generality, the feedback duration is assumed to be negligible due to its small data size and the edge server's high transmit power. Note that $\mathbb{S}^{(k+1)}$ can be available when the round $(k+1)$ starts, and a sufficient amount of time should be reserved for training the $(k+1)$-depth classifier. Denote $\tau_{k+1}$ as the corresponding training duration. Recalling the duration of one round being fixed to $t_0$, the offloading duration $t_{k+1}$ should be less than $t_0-\tau_{k+1}$, making energy consumption significant as $\tau_{k+1}$ becomes longer.

It can be overcome by offloading the partial data samples' $(k+1)$ features in advance during the training process, called  \emph{prefetching}. The resultant offloading duration can be extended from $t_{k+1}$ to $t_0$, enabling the {IoT device} to reduce energy consumption, according to \eqref{EQ : E}. On the other hand, the prefetching decision is based on predicting on whether the concerned data sample becomes ACSs. Unless correct, the excessive energy is consumed to prefetch useless features. Establishing this tradeoff is key, which will be addressed by formulating a stochastic optimization in Section. \ref{subsection:problem_formulation}.    

\subsection{Metrics of Clarity}\label{subsection:metric}

Consider an arbitrary round $k$, recalling that the $k$-depth classifier is trained by $\mathbf{x}_m^{(k)}$ for $ m \in \mathbb{S}^{(k)}$.
We formally define MoCs for binary SVM and multi-class DNN classifiers differently as follows\footnote{We unify all notations of MoCs as $\beta_m^{(k)}$ for simplicity since different MoC defintitions are not used at the same time.}.


\subsubsection{Binary SVM classifier}
A classic SVM whose $k$-depth decision hyperplane is given as
\begin{align}\label{Eq: Hyperplane}
    (\mathbf{w}^{(k)})^T\mathbf{x}^{(k)}+b^{(k)} =0, 
\end{align}
where $\mathbf{w}^{(k)} \in \mathbb{R}^k$ is the vector perpendicular to the hyperplane and $b^{(k)}$ is the offset parameter. The clarity of SVM can be measured by the distance to the hyperplane in \eqref{Eq: Hyperplane}. Given a data sample $\mathbf{x}_m^{(k)}$ for $m \in \mathbb{S}^{(k)}$, the distance-based MoC can be computed as
\begin{align}\label{eq:d_m^k}
    \beta_m^{(k)} = | (\mathbf{w}^{(k)})^T\mathbf{x}_m^{(k)} + b^{(k)} | / \|\mathbf{w}^{(k)}\|, 
\end{align}
where $\| \cdot \|$ represents the Euclidean norm. Then, the data sample $\mathbf{x}_m$ is included in a new ACS set $\mathbb{S}^{(k+1)}$ if $\beta_m^{(k)}$ is less than a threshold $\bar{\beta}^{(k)}$ to be specified in Section \ref{subsection:threshold_design}. \; It means that data samples near the hyperplane are more ambiguous and harder to classify correctly compared to those far from the hyperplane. On the other hand, the data samples in the CCS set can be predicted using the $k$-depth classifier as follows: 
\begin{align}
    \hat{c}_m^{(k)} = 
    \begin{cases}
        0 & \text{if } (\mathbf{w}^{(k)})^T\mathbf{x}_m^{(k)} + b^{(k)} \geq 0, \\
        1 & \text{if } (\mathbf{w}^{(k)})^T\mathbf{x}_m^{(k)} + b^{(k)} < 0,
    \end{cases} \quad  \forall m\in  \mathbb{S}^{(k)}\setminus \mathbb{S}^{(k+1)}.
\end{align}
where $\hat{c}_m^{(k)}$ is the predicted label of $\mathbf{x}_m^{(k)}$ by the $k$-depth classifier.

\subsubsection{Multi-class DNN classifier} \label{subsec:clarity_of_dnn}
Unlike the above distance-based MoC in the SVM classifier, a DNN-based classifier defines MoCs using a posterior distribution as follows. A DNN trained for the $k$-th classifier, parameterized by a large number of weights and bias functions, is collectively denoted as a function~$\boldsymbol{\theta}^{(k)}$ whose input and output are set to the raw data dimension $D$ and the number of classes $C$, respectively\footnote{Prompted by the fact that a DNN works well in high-dimensional input space, we fix the input dimension as $D$ regardless of the depth $k$. As recalled from Section. \ref{subsection:data_embedding}, the edge server knows the embedding function $\mathcal{F}$ of \eqref{Eq:Data_Embedding}. Then, the concerned data samples can be reconstructed in the original $D$-dimensional space using the received features from $1$ to $k$ while those from $(K+1)$ to $F$ are padded as zero.}. Given an arbitrary sample $\mathbf{x}_m^{(k)}$, we can compute the posterior likelihood of class $c$, denoted by  $P_{\boldsymbol{\theta}^{(k)}} (c | \mathbf{x}_m^{(k)})$, using a forward-propagation algorithm. 
Then, we can predict the sample's class via posterior maximization, given as
\begin{align}\label{eq:maximum_class}
    {c}_{m,\mathrm{1st}} = \mathop{\arg \max}\limits_{c\in\mathcal{C}} P_{\boldsymbol{\theta}^{(k)}}(c|\mathbf{x}_m^{(k)}), 
\end{align}
whose clarity depends on the distribution of posterior likelihoods across classes. 
We use the following two MoCs to this end.  
\begin{itemize}
    \item \textbf{Negative Entropy}: Entropy is widely used to quantify a given probabilistic distribution's uncertainty (see, e.g., \cite{SSEO_JSAC_2023}). For the purpose of quantifying the clarity, on the other hand, we use a negative entropy defined as
    \begin{align}
    \beta_m^{(k)} = \sum_{c\in \mathbb{C}} P_{\boldsymbol{\theta}^{(k)}}(c|\mathbf{x}_m^{(k)})\log P_{\boldsymbol{\theta}^{(k)}}(c|\mathbf{x}_m^{(k)}), 
    \end{align}
    which reaches its minimum when every posterior likelihood is equivalent.

    \item \textbf{Posterior Gap}: While the above negative entropy can be useful to measure the clarity averaged over entire classes, our main interest is in the specific class whose posterior likelihood is the maximum, as shown in \eqref{eq:maximum_class}. If the prediction in \eqref{eq:maximum_class} is wrong, then the class with the second largest likelihood is most likely correct. From this perspective, we use the gap of posterior likelihoods between the first and second largest classes, defined as
\begin{align}
    \beta_m^{(k)} = P_{\boldsymbol{\theta}^{(k)}}({c}_{m,\mathrm{1st}}|\mathbf{x}_m^{(k)}) - P_{\boldsymbol{\theta}^{(k)}}({c}_{m,\mathrm{2nd}}|\mathbf{x}_m^{(k)}),
    \end{align}
    where ${c}_{m,\mathrm{1st}}$ is specified in \eqref{eq:maximum_class} and ${c}_{m,\mathrm{2nd}}=\mathop{\arg \max}\limits_{c\in\mathcal{C}\setminus \{{c}_{m,\mathrm{1st}}\}} P_{\boldsymbol{\theta}^{(k)}}(c|\mathbf{x}_m^{(k)})$.
\end{itemize}
Similar to the above binary classifier, the data sample $\mathbf{x}_m$ is included in a new ACS set $\mathbb{S}^{(k+1)}$ if $\beta_m^{(k)}$ is less than a threshold $\bar{\beta}^{(k)}$. The other data samples, which are included in the CCS set, can be predicted as 
\begin{align}\label{eq:predict_label_dnn}
\hat{c}_m^{(k)} = {c}_{m,\mathrm{1st}}, \quad \forall m\in  \mathbb{S}^{(k)}\setminus \mathbb{S}^{(k+1)}.  
\end{align}

\subsection{Hierarchical Edge Inference}
After $K$ rounds, the entire classifier has a hierarchical structure comprising from $1$-depth to $K$-depth classifiers. Consider that a {IoT} device sends an unlabeled data sample to the edge server, which is initially set as an ACS. Starting from the $1$-depth classifier, the data sample passes through different depth classifiers in sequence until it is changed to an CCS. The last classifier's depth is referred to as the data sample's depth. In other words, its classification result becomes the final one.

\section{Threshold Design for Data Deepening}\label{subsection:threshold_design}
This subsection deals with the threshold design $\bar{\beta}^{(k)}$ to categorize whether the concerned data sample $\mathbf{x}_m^{(k)}$ is an ACS or a CCS based on the $k$-depth classifier. The thresholds targeting the SVM and DNN are introduced accordingly.

\subsection{Binary SVM Classifier}  

\begin{figure}[t] 
\centering
\centering
\includegraphics[width=8cm]{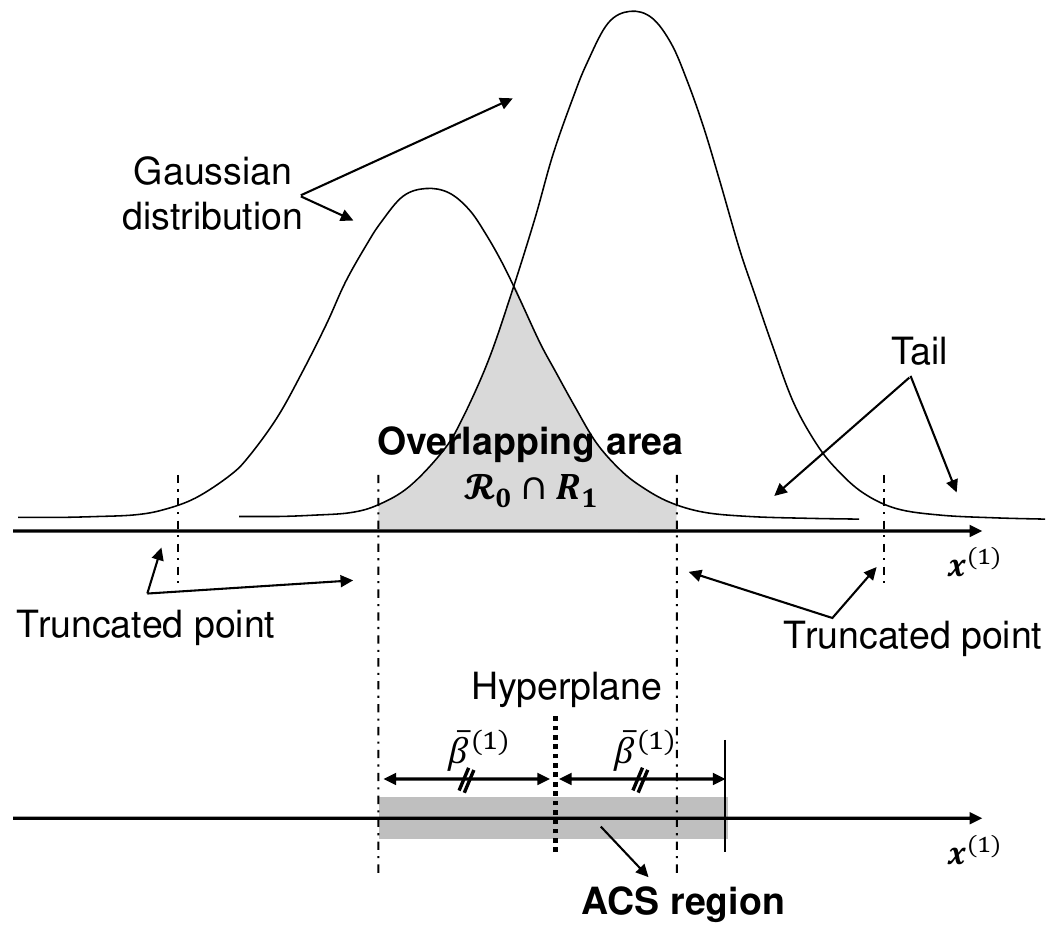}
\vspace{-10pt}
\caption{The ACS region in the 1-dimensional space is obtained by the probability distribution and distance from the hyperplane.}
\label{fig-gaussian}
\end{figure}

The stochastic distribution of each class can be approximated in the form of $k$-variate Gaussian processes using the \emph{Gaussian mixture model} (GMM) \cite{bishop2006pattern}. As shown in Fig. \ref{fig-gaussian}, the overlapping area between two distributions is observed. The data samples in the area are likely to be misclassified. We aim at setting the threshold $\bar{\beta}^{(k)}$ in such a way that most data samples in the overlapping area are included except a few outliers located in each tail.

To this end, we introduce the \emph{Mahalanobis distance} (MD) \cite{bensimhoun2009n} as a metric representing the distance from each instant to the concerned distribution. Given class $c\in\{0,1\}$, the MD is defined as
\begin{align}
    \delta_c^{(k)} = \sqrt{\left(\mathbf{x}^{(k)}-\boldsymbol{\mu}_c^{(k)}\right)^T\cdot \left(\boldsymbol{\Sigma}_c^{(k)}\right) ^{-1} \cdot \left(\mathbf{x}^{(k)}-\boldsymbol{\mu}_c^{(k)}\right)}, 
\end{align}
where $\mathbf{x}^{(k)}\sim \mathcal{N}(\boldsymbol{\mu}_c^{(k)}, \boldsymbol{\Sigma}_c^{(k)})$ with $\boldsymbol{\mu}_c^{(k)} \in \mathbb{R}^k$ and  $\mathbf{\Sigma}_c^{(k)} \in \mathbb{R}^{k\times k}$ being the distribution's mean vector and covariance matrix respectively, which are obtainable through the GMM process.
It is obvious that $\delta_c^{(k)}$ is a scale-free random variable and we attempt to set the threshold as the value whose \emph{cumulative distribution function} (CDF) of $\delta_c^{(k)}$ becomes $p_\mathrm{th}$, namely,
\begin{align}\label{eq:p_th(svm)}
\mathsf{Pr}\left[\delta_c^{(k)}\leq \bar{\delta}_c^{(k)}\right]=p_\mathrm{th}.
\end{align}
Noting that the square of $\delta_c^{(k)}$ follows a chi-square distribution with $k$ degrees-of-freedom, the CDF of this distribution for $r>0$ is defined as :
\begin{align}
    \mathscr{G}(r;k) = \mathsf{Pr} \left[ \delta_c^{(k)} \leq r\right] = \frac{\gamma\left(\frac{k}{2},\frac{r}{2}\right)}{\Gamma\left(\frac{k}{2}\right)},
\end{align}
where $\Gamma$ is the gamma function defined as $\Gamma(k) = \int_{0}^{\infty} t^{k-1}e^{-t}dt$ and $\gamma$ is the lower incomplete gamma function defined as $\gamma(k, r) = \int_{0}^{r} t^{k-1}e^{-t}dt$. The threshold $\bar{\delta}_c^{(k)}$ can be expressed as a closed-form as follows: 
\begin{align}\label{eq:delta_k}
    \bar{\delta}_c^{(k)} = \sqrt{\mathscr{G}^{-1}(p_\mathrm{th};k)},
\end{align}
where $\mathscr{G}^{-1}$ represents the inverse CDF of the chi-square distribution with $k$ degrees-of-freedom. Due to the scale-free property, the threshold $\bar{\delta}_c^{(k)}$ is identically set regardless of the concerned class; thus, the index of class can be omitted, namely,  $\bar{\delta}_0^{(k)}=\bar{\delta}_1^{(k)}=\bar{\delta}^{(k)}$. Given $\bar{\delta}^{(k)}$, each distribution can be truncated as 
\begin{align} \label{eq:R_c}
\mathcal{R}_c = \left\{ \mathbf{x}^{(k)}\in\mathbb{R}^{(k)} ~ | ~ \delta_c^{(k)} \leq \bar{\delta}^{(k)}\right\}, \quad c\in\{0,1\}.
\end{align}
{It can help to eliminate the outliers in each distribution.} Last, the threshold $\bar{\beta}^{(k)}$ is set by the maximum distance from the hyperplane in \eqref{Eq: Hyperplane} to an arbitrary $k$-dimensional point $\mathbf{x}^{(k)}$ in the overlapping area of $\mathcal{R}_0$ and $\mathcal{R}_1$, given as 
\begin{align} \label{eq:d_k}
    \bar{\beta}^{(k)} = \max_{\mathbf{x}^{(k)} \in \mathcal{R}_0 \cap \mathcal{R}_1} \left| (\mathbf{w}^{(k)})^T\mathbf{x}^{(k)} + b^{(k)} \right| / \|\mathbf{w}^{(k)}\|.
\end{align}
{The threshold $\bar{\beta}^{(k)}$ is determined using the Euclidean distance because the stochastic distribution is not considered when predicting data samples through the binary SVM classifier.} The process to obtain the threshold $\bar{\beta}^{(k)}$ is summarized in Algorithm \ref{algorithm:threshold}.

\begin{algorithm}[t]
\caption{Design of the threshold $\bar{\beta}^{(k)}$ based on the distance for SVM classifier}
\begin{algorithmic}[1]
    \Require Embedded data $\mathbf{x}_m^{(k)}$ for $m\in \mathbb{S}^{(k)}$, $k$-th version classifier.
    \For{$c \in \{0,1\}$}
        \State Find $\boldsymbol{\mu}_c$, $\boldsymbol{\Sigma}_c$ through GMM process.
        \State Compute $\bar{\delta}_c^{(k)}$ specified in \eqref{eq:delta_k}.
        \parState{%
        Compute the truncated domain of distribution $\mathcal{R}_c$ using \eqref{eq:R_c}}
    \EndFor
    \State Find the overlapping area $\mathcal{R} = \mathcal{R}_0 \bigcap \mathcal{R}_1$.
    \parState{%
    Using $k$-th version classifier in \eqref{Eq: Hyperplane}, compute {$\bar{\beta}^{(k)}$} using \eqref{eq:d_k}.}
    \\ \Return{$\bar{\beta}^{(k)}$}.\;
\end{algorithmic}
\label{algorithm:threshold}
\end{algorithm}

\begin{remark}[Symmetric ACS Region]\emph{Noting that each class's covariance matrix $\{\mathbf{\Sigma}_c^{(k)}\}_{c\in\{0,1\}}$ is different, the resultant truncated areas of $\mathcal{R}_0$ and $\mathcal{R}_1$ become asymmetric. To avoid the classifier being overfitted to one class, we choose the common distance threshold for both classes, say $\bar{\beta}^{(k)}$ in \eqref{eq:d_k}, corresponding to the maximum distance between the two.}
\end{remark}

\subsection{Multi-Class DNN Classifier}
Contrary to the above SVM classifier, it is challenging to approximate the posterior distribution for each class of the DNN classifier. Instead, the error distribution can be empirically computed by normalizing the number of prediction errors, namely, $\hat{c}_m^{(k)}\neq {c}_m$, where $\hat{c}_m^{(k)}$ specified in \eqref{eq:predict_label_dnn} represents the predicted label of the data sample $\mathbf{x}_m$ using the $k$-th depth classifier $\boldsymbol{\theta}^{(k)}$ and ${c}_m$ is the corresponding ground truth.
Specifically, the joint distribution of the prediction error and MoC $\beta_m^{(k)}$, denoted by $z^{(k)}(\beta) =\Pr[\hat{c}_m \neq c_m ~,~\beta_m^{(k)} \geq \beta]$, is given as
\begin{align}\label{eq:joint_distribution}
    z^{(k)}(\beta) = \frac{\sum_{m\in\mathbb{S}^{(k)}}\mathsf{I}(\hat{c}_m \neq c_m ~,~\beta_m^{(k)} \geq \beta)}{|\mathbb{S}^{(k)}|},
\end{align}
where $\mathsf{I}(x)$ is an indicator function returning $1$ if $x$ is true and $0$ otherwise. Recalling the definition of MoC explained in Section. \ref{subsec:clarity_of_dnn}, it is straightforward that the joint distribution $z^{(k)}(\beta)$ is a monotone non-increasing function with respect to $\beta$. Given an allowable error probability denoted by $z_{\mathrm{th}} \in [0,1]$, we can find the threshold $\bar{\beta}^{(k)}$ by solving the following one-dimensional searching problem: 
\begin{align}\label{eq:p_th(dnn)}
    \bar{\beta}^{(k)} = \sup\{\beta ~|~ z^{(k)}(\beta) \geq z_{\mathrm{th}}\}.
\end{align}

\begin{remark}  [Communication Efficiency vs. Learning Accuracy] \label{remark:parameter} \emph{Due to the nature of a random data distribution, a few errors are inevitable for both binary SVM and multi-class DNN classifiers. The parameters $p_{\mathrm{th}}$ and $z_{\mathrm{th}}$ determine their allowable levels, allowing us to balance between communication efficiency and learning accuracy. Specifically, when these parameters increase, fewer data samples are included in $\mathbb{S}^{(k+1)}$ as specified in \eqref{eq : ACI set}, resulting in sending the $(k+1)$-th features of less data samples while the learning accuracy can be degraded. We verify this tradeoff relation in terms of $p_{\mathrm{th}}$ and $z_{\mathrm{th}}$ by extensive simulation studies in Section. \ref{subsec:tradeoff-deepening}.}
\end{remark}

\section{Optimal Data Prefetching}\label{section:optimal_data_prefetching}
This section deals with selecting the size of prefetched data in the sense of  minimizing the expected energy consumption at the {IoT device}.

\subsection{Problem Formulation}
\label{subsection:problem_formulation}

Consider the prefetching duration in round $k$, say $\tau_k$, which is equivalent to the training duration of the $k$-depth classifier, as shown in Fig. \ref{Fig-prefetching}. The number of data samples in $\mathbb{S}^{(k)}$ is denoted by $s_k$. Among them, $p_k$ data samples are randomly selected and their $(k+1)$-th features are prefetched.  The prefetched data size is $\alpha p_k$, where $\alpha$ represents the number of bits required to quantize feature data\footnote{The quantization bit rate depends on the value of intensity. For example, one pixel in MNIST data has $255$ intensities and can be quantized to 8 bits, i.e., $\alpha = 8$.}. Given the channel gain $h_{k}$, the resultant energy consumption for prefetching is 
\begin{align} \label{EQ :energy(p_k)}
    \mathbf{E}(\alpha p_k, \tau_k ; h_k) = 
    \lambda \frac{\alpha^\ell }{h_{k}\tau_k^{\ell-1}}p_k^\ell,
\end{align}
where the monomial order $\ell$ is specified in Section. \ref{section:system model}. Here, the number of prefetched data $p_k$ is a discrete control parameter ranging from $0$ to $ s_k$. For tractable optimization in the sequel, we regard $p_k$ as a continuous variable within the range, which is rounded to the nearest integer in practice.   

Next, consider the offloading duration in round $(k+1)$, say $t_{k+1}=t_0-\tau_{k+1}$. Among the data samples in $\mathbb{S}^{(k+1)}$, a few number of data, denoted by $n_{k+1}$, remain after the prefetching. Given the channel gain $h_{k+1}$, the resultant energy consumption is 
\begin{align}
    \mathbf{E}(\alpha n_{k+1}, t_{k+1}; h_{k+1}) = \lambda \frac{\alpha^\ell }{h_{k+1}t_{k+1}^{\ell-1}}n_{k+1}^\ell.
\end{align}
Note that $n_{k+1}$ is determined after the $k$-depth classifier is trained. In other words, $n_{k+1}$ is random at the instant of the prefetching decision. Denote $\rho_k$ as the ratio of a data sample in $\mathbb{S}^{(k)}$ being included in $\mathbb{S}^{(k+1)}$ {and refer to it as the reduction ratio\footnote{{The reduction ratio $\rho_k$ depends on the characteristics of the dataset. It is assumed that the edge server knows $\rho_k$ in advance from historical data.  
}}.} Then, $n_{k+1}$ follows a binomial distribution with parameters $(s_k-p_k)$ and $\rho_k$, whose probability mass function is $P(j)=\bigl(\genfrac{}{}{0pt}{}{s_k-p_k}{j}\bigr)\rho_k^j (1-\rho_k)^{s_k-p_k-j}$ for $j=0,\cdots, {s_k-p_k}$. Given $p_k$, the expected energy consumption~is 
\begin{align}
&\mathbb{E}_{n_{k+1}, h_{k+1}}[\mathbf{E}(\alpha n_{k+1}, t_{k+1}; h_{k+1})]
=\lambda\frac{\nu\alpha^\ell}{t_{k+1}^{\ell-1}}\mathbb{E}_{n_{k+1}}[n_{k+1}^\ell],\label{EQ : energy(n_k)}
\end{align}
where $\nu=\mathbb{E}[\frac{1}{h_{k+1}}]$ is the expectation of the inverse channel gain, which can be known a priori due to its i.i.d. property.   

Last, summing up \eqref{EQ :energy(p_k)} and \eqref{EQ : energy(n_k)} corresponds to the expected energy consumption for the $(k+1)$-th feature when $p_k$ data samples are prefetched, leading to the following two-stage stochastic optimization: 
\begin{align}
    \min_{p_k} ~ & \lambda \frac{\alpha^\ell }{h_{k}\tau_k^{\ell-1}}p_k^\ell + \lambda\frac{\nu\alpha^\ell}{t_{k+1}^{\ell-1}}\mathbb{E}_{n_{k+1}}[n_{k+1}^\ell]  \nonumber\\
    \text{s.t.} \quad & 0 \leq p_k \leq s_k. \nonumber
    \label{problem:overall} \tag{P1} \nonumber
\end{align}
The optimal prefetching policy can be designed by solving  \ref{problem:overall}, which enables us to balance the energy consumption between the dissipation due to offloading data samples not used in the next round and the savings by extending the available transfer time. The detailed derivation will be presented in the following subsection. 

\subsection{Optimal Prefetching Control}
This subsection aims at deriving the closed-form expression of the optimal prefetching number $p_k^*$ by solving \ref{problem:overall}. The main difficulty lies in addressing the $\ell$-th moment $\mathbb{E}_{n_{k+1}}[n_{k+1}^\ell]$, of which the simple form is unknown for general $\ell$. To address it, we refer to the upper bound on the $\ell$-th moment in \cite{2ahle2022sharp},
\begin{align}
    \mathbb{E}_{n_{k+1}}[n_{k+1}^{\ell}] \leq \left(\mu_{n_{k+1}} + \frac{\ell}{2}\right)^\ell,
\end{align}
where $\mu_{n_{k+1}} = \left(s_k-p_k\right)\rho_k$ is the mean of the binomial distribution with parameters $(s_k-p_k)$ and $\rho_k$. It is proved in \cite{2ahle2022sharp} that the above upper bound is tight when the order $\ell$ is less than the mean $\mu_{n_{k+1}}$. Therefore, instead of solving \ref{problem:overall} directly, we alternatively formulate the problem of minimizing the upper bound as 
\begin{align}
    \min_{p_k} ~& \frac{p_k^\ell}{h_{k}\tau_k^{\ell-1}} + \frac{\nu}{t_{k+1}^{\ell-1}} \left( \left(s_k-p_k\right)\rho_k + \frac{\ell}{2}\right)^\ell
    \nonumber\\
    \text{s.t.} \quad & 0 \leq p_k \leq s_k. \nonumber \tag{P2}\label{Problem:upper bound}
\end{align}
Note that \ref{Problem:upper bound} is a convex optimization, enabling us to derive the closed-form solution. The main result is shown in the following proposition. 
\begin{proposition}[Optimal Prefetching Policy]\label{proposition 1}\emph{Given the ratio of prefetching $\rho_k$ in round $k$, the optimal prefetching data size $p_k^*$, which is the solution to \ref{Problem:upper bound}, is
\begin{align}\label{eq:Optimal_sol}
    p^*_k = \left(\frac{\varphi\rho_k^{\frac{1}{\ell-1}}}{1+\varphi\rho_k^{\frac{\ell}{\ell-1}}}\right) \left(s_k \rho_k + \frac{\ell}{2}\right),
\end{align}
where $\varphi = \left(h_k\nu \right)^{\frac{1}{\ell-1}}\frac{\tau_k}{t_{k+1}}$ and $s_k$ is the number of data samples in $\mathbb{S}^{(k)}$.}
\end{proposition}
\begin{IEEEproof}
See Appendix. \ref{appendix1}.
\end{IEEEproof}

\begin{figure}[t] 
\centering
\centering
\includegraphics[width=8.1cm]{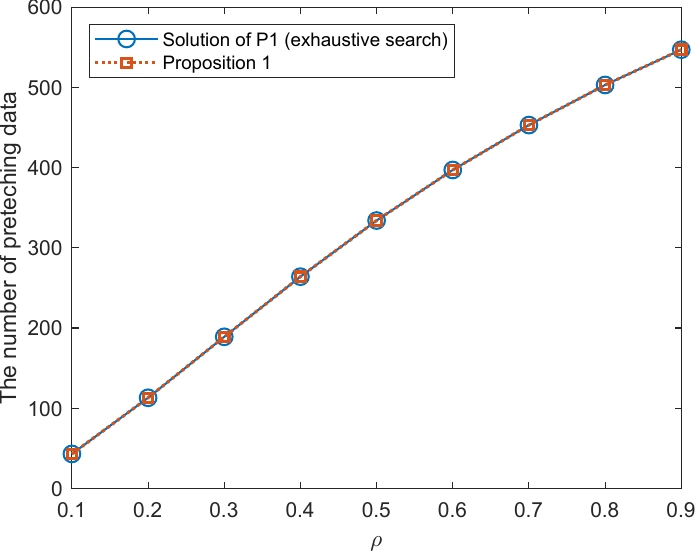}
\vspace{-5pt}
\caption{{The number of prefetching data obtained by Proposition \ref{proposition 1} is equal to the number of optimal prefetching data. Parameters are set $s_k = 1000$, $\ell = 3$ and $\tau_k = t_{k+1} = 0.5$.}}
\label{fig:optimality}
\end{figure}

{As shown in Fig. \ref{fig:optimality}, the number of prefetched data in Proposition \ref{proposition 1} is shown to be near-to-optimal with the comparison of the optimal one obtained through an exhaustive search.}

\begin{remark}[Effect of Parameters]\emph{When assuming that the number of ACSs in slot $k$, say $s_k=|\mathbb{S}^{k}|$, is significantly larger than $\frac{\ell}{2}$, we can approximate \eqref{eq:Optimal_sol} as  $p^*_k \approx \left(\frac{\varphi\rho_k^{\frac{1}{\ell-1}}}{1+\varphi\rho_k^{\frac{\ell}{\ell-1}}}\right) s_k \rho_k$. Noting that the term $s_k\rho_k$ represents the expected number of ACSs in slot $(k+1)$, the parameter $\varphi$ controls the portion of prefetching as follows:   
\begin{itemize}
    \item As the current channel gain $g_k$ becomes larger or the training duration $\tau_k$ increases, the parameter $\varphi$ increases and the optimal solution $p_k^*$ reaches near to the $s_k\rho_k$;   
    \item As $g_k$ becomes smaller and $\tau_k$ decreases, both $\varphi$ and $p_k^*$ converge to zero.  
\end{itemize}}
\end{remark}

\begin{figure*}[t]
\centering
\includegraphics[width=14cm]{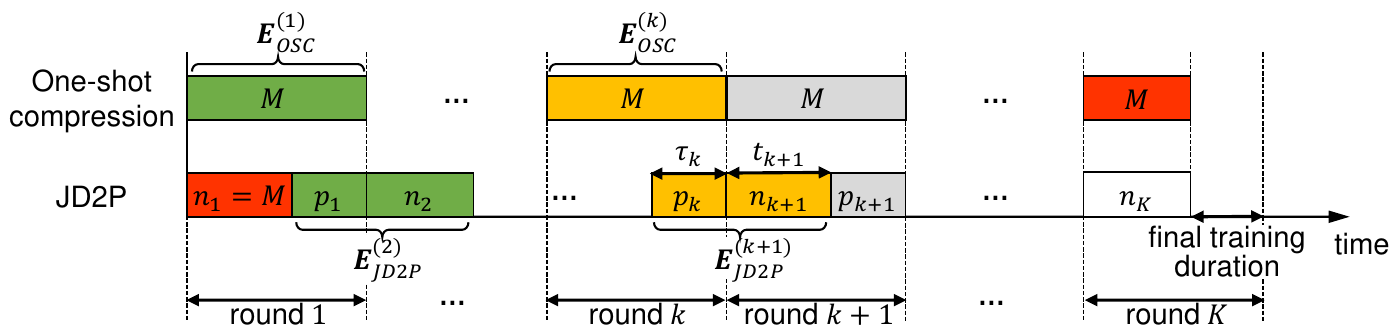}
\vspace{-5pt}
\caption{The number of data samples being offloaded in each duration when OSC and JD2P are applied respectively. Each area with the same color is compared with each other. The total duration of each round is $t_0$.}  
\label{Fig-energy_efficiency}
\end{figure*}

\subsection{Energy Efficiency Analysis}\label{subsection:energy.efficiency.analysis}

This subsection aims to investigate the proposed J2DP's energy efficiency by comparing it with the method offloading every data sample's features from $1$ to $K$, referred to as an \emph{one-shot compression} (OSC) throughout the paper.

As illustrated in Fig. \ref{Fig-energy_efficiency}, OSC does not require data prefetching since there is only one training process after finishing data offloading. Its expected energy consumption, denoted as $\mathbf{E}_{\mathrm{OSC}}$, can be derived using \eqref{EQ : E}:
\begin{align}
    \mathbf{E}_{\mathrm{OSC}} & = \mathbb{E}_{h}\left[ \sum_{k=1}^{K-1} \mathbf{E}(\alpha M, t_0;h_k) + \mathbf{E}(\alpha M, t_K;h_K)\right] \nonumber\\
    & = \mathbb{E}_h \left[ \sum_{k=1}^{K-1} \lambda \frac{(\alpha M)^\ell}{h_k {t_0}^{\ell-1}} + \lambda \frac{(\alpha M)^\ell}{h_K {t_K}^{\ell-1}}\right]\nonumber\\
    & = (K-1) \underbrace{\lambda \nu \frac{(\alpha M)^{\ell}}{{t_0}^{\ell-1}}}_{\triangleq~ \mathbf{E}_{\mathrm{OSC}}^{(k)}} + \lambda \nu \frac{(\alpha M)^\ell}{{t_K}^{\ell-1}}, \label{eq:E_bench}
\end{align}
where $\nu$ represents the expectation of the inverse channel gain specified in \eqref{EQ : energy(n_k)}, $M$ is the total number of data samples, and $t_0$ is the duration of a round. On the other hand, the total expected energy consumption of JD2P, denoted by $\mathbf{E}_{\mathrm{JD2P}}$, can be derived as
\begin{align}
    & \mathbf{E}_{\mathrm{JD2P}}  \nonumber \\
    & = \mathbb{E}_{h} \bigg[ \mathbf{E}(\alpha n_1, t_1;h_1) +  \nonumber \\ 
    & \quad \quad \sum_{k=2}^{K} \left(\mathbf{E}(\alpha p^*_{k-1}, \tau_{k-1};h_{k-1}) + \mathbf{E}(\alpha n_{k}, t_{k};h_{k})\right)\bigg] \\
    & = \mathbb{E}_{h} \bigg[ \lambda \frac{(\alpha n_1)^{\ell}}{h_{1} {t_1}^{\ell-1}} \nonumber \\ 
    & \quad \quad \quad + \sum_{k=2}^{K} \lambda \left(\frac{(\alpha p^*_{k-1})^{\ell}}{h_{k-1} {\tau_{k-1}}^{\ell-1}} + \mathbb{E}_{n_{k}}\left[\frac{(\alpha n_{k})^{\ell}}{h_{k}{t_{k}}^{\ell-1}}\right]\right)\bigg] \nonumber\\
    & = \lambda \nu \frac{(\alpha n_1)^{\ell}}{{t_1}^{\ell-1}} + \sum_{k=2}^{K} \underbrace{ \lambda \nu \left( \frac{(\alpha p^*_{k-1})^{\ell}}{{\tau_{k-1}}^{\ell-1}} + \mathbb{E}_{n_{k}}\left[\frac{(\alpha n_{k})^{\ell}}{{t_{k}}^{\ell-1}} \right]\right)}_{\triangleq~\mathbf{E}_{\mathrm{JD2P}}^{(k)}}, \label{eq:E_jd2p}
\end{align}
where $n_k$ is the number of data samples whose $k$-th feature is offloaded in round $k$. It is obvious that the first feature of every data sample should be offloaded, namely, $n_1=M$. We assume that $t_k$ is equivalent for all $k$'s without loss of generality, yielding that the last term in \eqref{eq:E_bench} and the first term in \eqref{eq:E_jd2p} is equivalent. In other words, JD2P's energy efficiency can be shown through a pairwise energy-consumption comparison between $\mathbf{E}_{\mathrm{OSC}}^{(k)}$ and $\mathbf{E}_{\mathrm{JD2P}}^{(k+1)}$ in \eqref{eq:E_bench} and \eqref{eq:E_jd2p}, respectively, namely, 
\begin{align}\label{eq:energy_efficiency} 
\frac{\mathbf{E}_{\mathrm{JD2P}}^{(k+1)}}{\mathbf{E}_{\mathrm{OSC}}^{(k)}} <1, \quad 1 \leq k \leq K-1.
\end{align}
J2DP is said to be more energy-efficient than OSC if the above condition for every $k\in\{1, \cdots , K-1\}$ is satisfied. For $\mathbf{E}_{\mathrm{JD2P}}^{(k+1)}$, the number of the optimal prefetching data samples, say $p_{k}^*$ of \eqref{eq:Optimal_sol}, is shown to be upper bound by $s_{k}\rho_{k}$. With the upper bound, the number of the remaining data samples is thus $s_{k}-s_{k}\rho_{k}$, and the number of offloaded data samples among them follows a binomial distribution with parameters $s_{k}-s_{k}\rho_{k}$ and $\rho_{k}$. Using the $\ell$-th moment of a binomial distribution, the expected energy consumption $\mathbf{E}_{\mathrm{JD2P}}^{(k+1)}$ can be upper bounded by
\begin{align}\label{eq:upperbound_for_Jd2p}
    \mathbf{E}_{\mathrm{JD2P}}^{(k+1)} \leq \lambda \nu \alpha^{\ell}\left(\frac{(s_{k} \rho_{k})^\ell}{{\tau_{k}}^{\ell-1}} + \frac{(s_{k}\rho_{k} (1-\rho_{k}))^{\ell}}{{t_{k+1}}^{\ell-1}}\right).
\end{align}
By plugging the above into \eqref{eq:energy_efficiency}, we have
\begin{align}
& \frac{\lambda \nu \alpha^{\ell}\left(\frac{1}{{\tau_{k}}^{\ell-1}}+\frac{(1-\rho_{k})^{\ell}}{{t_{k+1}}^{\ell-1}}\right)(s_{k} \rho_{k})^{\ell}}{\lambda \nu \frac{(\alpha M)^ {\ell}}{{t_0}^{\ell-1}}}  \\
& = \left(\left(\frac{1}{\gamma}\right)^{\ell-1}+\left(\frac{1}{1-\gamma}\right)^{\ell-1}(1-\rho_{k})^{\ell}\right){\rho_{k}^\ell},\label{eq:energy_gain}
\end{align}
where $\gamma = \frac{\tau_{k}}{t_0}$ denotes the portion of the training duration. Assume $\rho_{k}$ to be near to $1$. One observes that, as $\rho$ becomes closer to one, the second term is a dominant factor, allowing us to approximate \eqref{eq:energy_gain} as $\left(\frac{1}{1-\gamma}\right)^{\ell-1}(1-\rho_{k})^{\ell} \rho_{k}^\ell$. Thus, the condition of energy efficiency is given as
\begin{align} \label{eq:condition}
    \frac{\tau_{k}}{t_0} < 1 - \left( \rho_{k} (1- \rho_{k}) \right)^{\frac{\ell}{\ell-1}}.
\end{align}

\begin{remark} [Effect of Parameters]
    \emph{Recalling $\rho_{k} \in [0,1]$ gives that the value $\rho_{k}(1-\rho_{k-1})$ in \eqref{eq:condition} is upper-bound by $\frac{1}{4}$. Then, the right-hand side of the condition is reduced to $0.93$ and $0.98$ when $\ell$ is set to $2$ and $3$, respectively. That is to say, the JD2P is said to be energy-efficient in most cases unless the training duration $\tau_{k}$ is extremely large.}
\end{remark}

\begin{remark}[Channel Capacity vs. Energy Consumption]\emph{{The expected energy consumption of JD2P derived in \eqref{eq:E_jd2p} is upper-bounded by \eqref{eq:upperbound_for_Jd2p}, showing that the expected energy consumption is proportional to the expectation of the inverse channel gain, denoted as $\nu$ which affects the channel capacity. In other words, the expected energy consumption depends not only on the expected channel gain but also on its variance. For example, although the expected channel gain is the same, the expected energy consumption can increase with high channel fluctuations. It is demonstrated in the simulation study in Sec. \ref{subsec:effec_of_channel}. 
}}    
\end{remark}

\section{Extension to General Cases}\label{section:extension_to_general_cases}
{
The JD2P algorithm is proposed based on centralized edge learning with historical data and training data given in advance, considered offline machine learning. The pair of single IoT devices and edge servers are only considered as the system model. The proposed algorithm can be extended to two major research topics. One is an online learning system that has many attractions for large-scale machine learning tasks in real-world applications \cite{hoi2021online}. The other is a multi-user federation edge learning system. The discussion of extensions is introduced as follows.}

{
Online learning aims to update an AI model in sequential and continuous manners, which can be broadly categorized into two approaches. First, when an AI model is gradually updated by dividing a large dataset into smaller ones \cite{xu2018new}, it is straightforward to apply data deepening such that an approach based on feature importance can be utilized to determine how the dataset should be partitioned. On the other hand, as the second approach, when updating the model through data arriving in real-time \cite{gomes2019machine}, the technique of data prefetching can be utilized to transmit data in advance while the classifier is being updated. At this time, leveraging previously trained classifiers to assess data depth allows for efficient resource allocation and the selection of relevant data. This allows us to enhance the high utilization of radio resources for training.
}

{
When applying J2DP into multi-user federated edge learning, one key factor that we should consider is that each IoT device has a heterogeneous dataset. Depending on the depth of the given data samples, the deepening level required at each IoT device can be different, bringing about the diverge of the classier model, as in the case of \cite{sattler2020clustered}. It can be overcome by clustering devices according to the data depth or importance, and IoT devices in the equivalent cluster can cooperate to train a classifier. Besides, the scheduling policy to prioritize clusters should be jointly considered due to the limited radio resources.
}


\section{Simulation Results}\label{Section:Simulation}
This section presents simulation results to validate the proposed JD2P's effectiveness over several benchmarks. The parameters are set as follows unless stated otherwise.

\subsection{Simulation Setting}
\subsubsection{Offloading Model}\label{subsubsec:setting_offloading}
The entire offloading duration consists of $10$ rounds (i.e., $K=10$), each of which is fixed to $t_0 = 0.1$ (sec). For offloading, the channel follows the gamma distribution with the shape parameter $\beta>1$ and the probability density function $f_h(x) = \frac{x^{\beta-1}e^{-\beta x}}{(1/\beta)^\beta\Gamma(\beta)}$, where the gamma function $\Gamma(\beta) = \int_0^\infty x^{\beta-1}e^{-x}dx$ and the mean $\mathbb{E}[h_k]=1$. The parameter is set as $\beta = 2$. The energy coefficient $\lambda$ is set to $10^{-17}$ according to \cite{tao2019stochastic}. The monomial order of the energy consumption model in \eqref{EQ : E} is set as $\ell =3$. 

For computing and prefetching, the reserved training duration $\tau_k$ is assumed to be constant and fixed to $\tau_k = \tau =0.5$ (sec) for $1\leq k\leq K$. On the other hand, prefetching is not required in round $K$, and the data offloading is only allowed for $t_{K} = t_0 - \tau$ (sec). 

\subsubsection{Dataset}\label{subsubsec:classifier_training}
We use the well-known MNIST dataset, including $M=6 \cdot 10^4$ training samples and $10^4$ testing samples. Each sample consists of $784$ gray-scaled pixels and is associated with one of $10$ classes. For training binary SVM classifiers, we conduct experiments with every possible pair of classes, namely, $\binom{10}{2}=45$ pairs. {Other datasets can be also considered, and the results of using Fashion MNIST are shown in Appendix \ref{appendix2}.}

The edge server utilizes the received data samples for two different training strategies, summarized as follows.
\begin{itemize}
    \item \textbf{Strategy 1}: This approach only utilizes the data samples in $\mathbb{S}^{(k)}$ for training the $k$-depth classifier. In other words, the number of data samples that participate in the training gradually decreases as the round goes by.  
    \item \textbf{Strategy 2}: For training the $k$-depth classifier, the server utilizes all received data samples so far, namely $\bigcup_{i=1}^k  \{\mathbf{x}_{m,i}^{(i)}\}_{m\in \mathbb{S}^{(i)}}$. For training the binary SVM classifier, data samples with fewer than $k$ features, namely $\mathbf{x}_m \notin \mathbb{S}^{(k)}$, are augmented into a $k$-dimension space with zero-padding. For training the DNN classifier, data samples are reconstructed into the dimension $D$ in every round, specified in Sec. \ref{subsection:metric}.
\end{itemize}
{The second strategy utilizes more data than the first one for training due to using all the received data.} In other words, the resultant learning accuracy of the latter is expected to be better than the former while paying the cost of computation complexity.

\subsubsection{Classifier Model}
As previously mentioned, we considered two different classifier models as $k$-depth classifiers: SVM and DNN. For the binary SVM, we used a soft-margin SVM with a slack variable set to $1$, which was implemented using the scikit-learn library in Python. For the DNN model, we used a $6$-layer network trained with a stochastic gradient descent algorithm and activated by a rectified linear unit function. During the training process, we set the mini-batch size to $64$ and the number of epochs to $10$ in every round. The DNN training was implemented using the PyTorch library with Nvidia RTX 2080 Ti GPU.

\subsubsection{Benchmarks}
We consider the following three benchmarks. The first one is OSC as explained in Section. \ref{subsection:energy.efficiency.analysis}, where all data samples' features from $1$ to $K$ are utilized to train a classifier. The second one is random data sampling where a finite number of data samples are randomly offloaded to train a classifier. The third one is random feature sampling where a finite number of data samples are randomly selected and their corresponding features are offloaded to train a classifier. 
The fourth one is motivated by \cite{liu2020wireless}, where a few number of data samples are selected according to their importance measured by following the methods therein, referred to as importance-aware data sampling. The amounts of offloading data for the second, third, and fourth benchmarks are equivalent to that of JD2P for fair comparisons.

\subsection{Data Deepening Ratio vs. Learning Accuracy}\label{subsec:tradeoff-deepening}

\begin{figure}[t] 
\centering
\centering
\subfigure[SVM classifier]{\includegraphics[width=8cm]{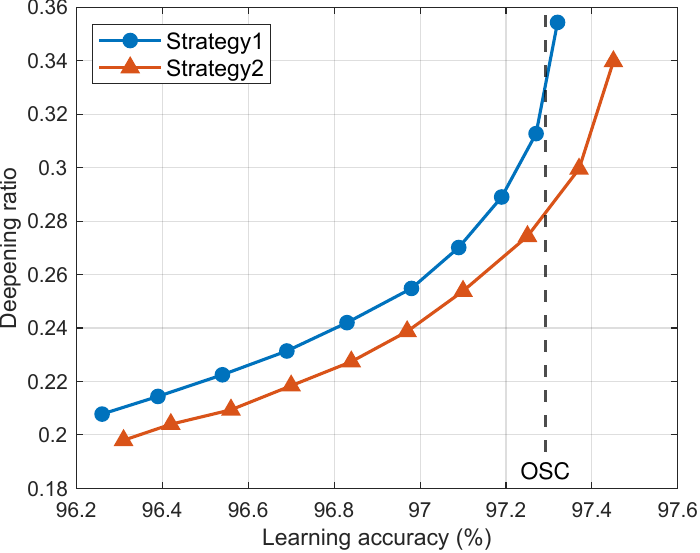}\label{fig:tradeoff_svm}}
\subfigure[DNN classifier]{\includegraphics[width=8cm]{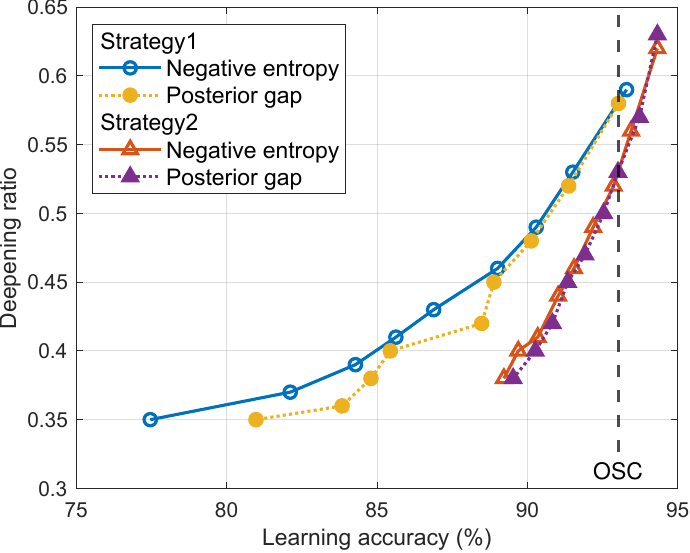}\label{fig:tradeoff_dnn}}
\caption{Data deepening exhibits a tradeoff between the learning accuracy and the deepening ratio. The performance is evaluated using the MNIST dataset. The deepening ratio is obtained by comparing the amount of data samples with the OSC. We consider (a) the SVM classifier ($p_{\mathrm{th}} \in [0.95, 0.995]$) and (b) the DNN classifier ($z_{\mathrm{th}} \in [0.01,0.09]$).}
\label{fig:tradeoff}
\vspace{-10pt}
\end{figure}

First, we evaluate the effect of data deepening on learning accuracy. We define a deepening ratio as the amount of training data with data deepening normalized by that of OSC, namely,   
\begin{align}
    \text{deepening ratio}  
    =  \frac{\sum_{k=1}^K|\mathbb{S}^{(k)}|}{K \times M}. 
\end{align}
Fig. \ref{fig:tradeoff} represents the tradeoff relation between the deepening ratio versus the learning accuracy. To this end, we control the parameters $p_\mathrm{th}$ in \eqref{eq:p_th(svm)} and $z_\mathrm{th}$ in \eqref{eq:p_th(dnn)}, which represent allowable error levels when the binary SVM classifier and the multi-class DNN classifier are used, respectively, as explained in Remark \ref{remark:parameter}. The dotted vertical line represents the learning performance of OSC. Several interesting observations are made. First, the proposed data deepening utilizes fewer data samples than those of OSC without considerable performance degradation. With the deepening ratio larger than a certain value, its learning accuracy is rather better than the benchmark since our data deepening can exclude several outliers hampering the learning performance, {which coincides with the existing literature such as \cite{janecek2008relationship} and \cite{golinko2019generalized}.} Second, with the same level of the deepening ratio, strategy $2$ is superior to strategy $1$ for both binary and multi-class classifiers as expected in Section. \ref{subsubsec:classifier_training}. Third, for a multi-class DNN classifier in Fig. \ref{fig:tradeoff_dnn}, we compare two MoCs specified in Section. \ref{subsec:clarity_of_dnn}, showing that the MoC based on the posterior gap always provides a better learning accuracy than that of the negative entropy-based one. The second and third observations guide us to use strategy $2$ and the MoC based on the posterior gap as the primary options of data deepening in the following simulation studies.  

\begin{figure}[t] 
\centering
\centering
\subfigure[SVM classifier]{\includegraphics[width=8cm]{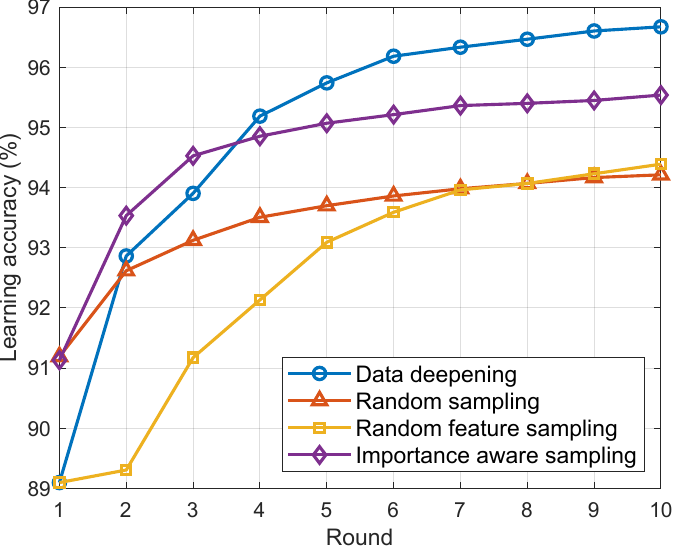}\label{fig:compare_svm}}
\subfigure[DNN classifier]{\includegraphics[width=8cm]{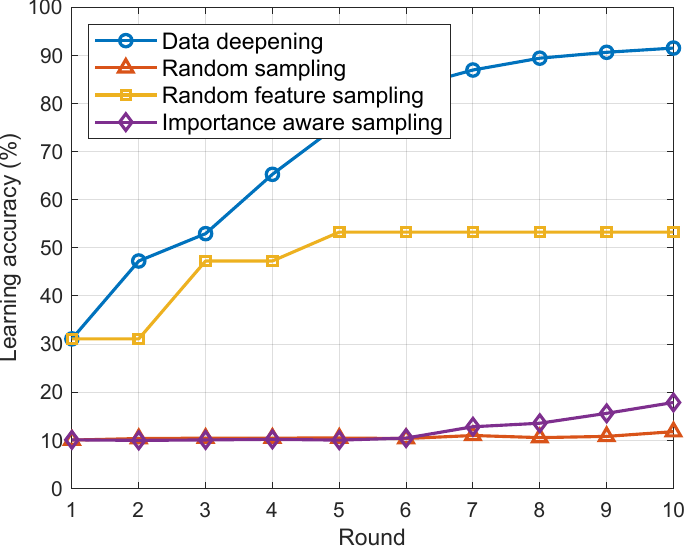}\label{fig:compare_dnn}}
\vspace{-5pt}
\caption{The learning performance of data deepening is compared with several benchmarks. Benchmarks use the same number of data samples that data deepening uses. We consider (a) the SVM classifier ($p_{\mathrm{th}} \in [0.95, 0.995]$) and (b) the DNN classifier ($z_{\mathrm{th}} \in [0.01,0.09]$).}
\label{fig:compare}
\vspace{-10pt}
\end{figure} 

\subsection{Classifier Depth vs. Learning Accuracy}

\begin{figure}[t] 
\centering
\centering
\subfigure[SVM classifier]{\includegraphics[width=8.2cm]{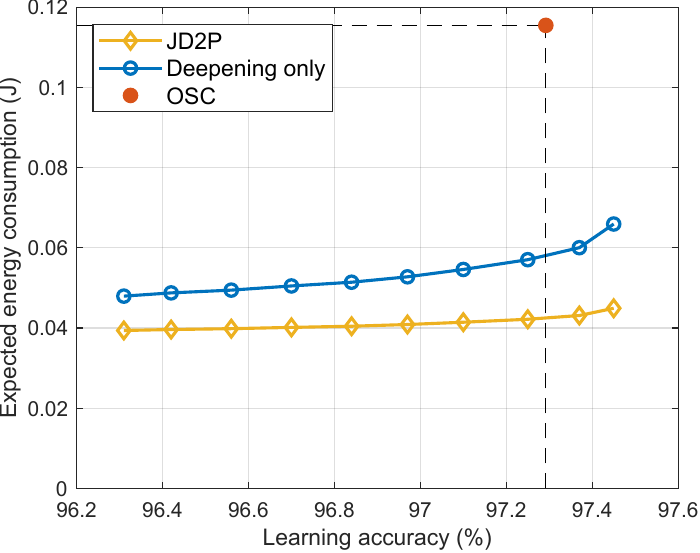}\label{fig:energy_tradeoff_svm}}
\subfigure[DNN classifier]{\includegraphics[width=8.2cm]{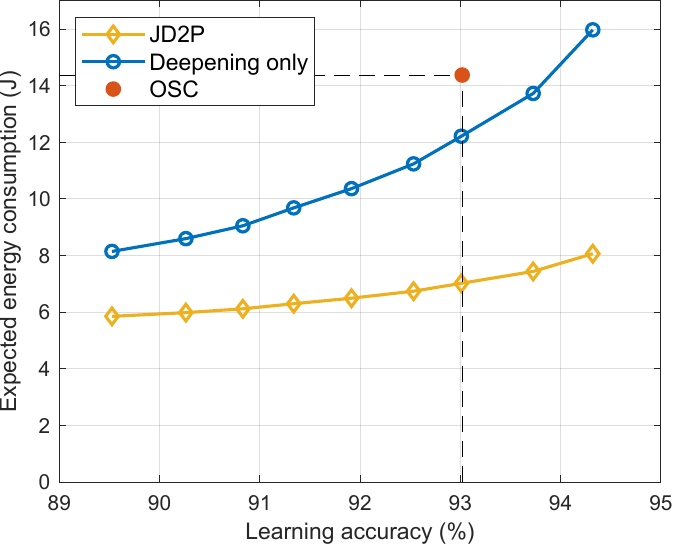}\label{fig:energy_tradeoff_dnn}}
\vspace{-3pt}
\caption{JD2P exhibits a tradeoff between the learning accuracy and the expected energy consumption. Deepening only represents the case without prefetching in JD2P. The OSC is compared as the benchmark. Each point represents the average value over error-tolerance parameters with $p_{\mathrm{th}} \in [0.95, 0.995]$ and $z_{\mathrm{th}} \in [0.01,0.09]$. The training duration is set as $\tau = 0.5$s.}
\label{fig:energy_tradeoff}
\end{figure}

Fig. \ref{fig:compare} shows the tendency of learning accuracy regarding classifier depths, equivalent to the number of rounds $K$. J2DP and all benchmarks tend to improve learning performance, while several observations are made differently between binary SVM and multi-class DNN classifiers. For the binary SVM classifier, the proposed data deepening provides a worse accuracy than random and importance-aware sampling benchmarks when the concerned number of rounds is relatively small, say $3$. In other words, the data embedding result with less than $3$ features is insufficient to achieve an acceptable accuracy. On the other hand, as more features are involved, the proposed data deepening outperforms the other benchmarks. Along with the raw data's dimension $D=784$, the resultant accuracy is significantly high with a small number of features, say nearly $97\%$, when the depth is $10$. Besides, the significant gap in the random feature sampling verifies the proposed data deepening's effectiveness on learning performance.

Next, for the multi-class DNN classifier illustrated in Fig. \ref{fig:compare_dnn}, it is observed that the data deepening outperforms all benchmarks for all depths. Recall that the amounts of offloading data samples for random and importance-aware sampling approaches are equivalent to the data deepening. Each data sample consists of $784$ pixels, and only a small number of data samples can be offloaded for the benchmarks. Since a DNN's performance depends on the number of data samples, the resultant accuracies are worse than expected. On the other hand, data deepening allows the edge server to utilize a sufficient number of data samples reconstructed through a finite number of features. It is shown that more than $90\%$ of the learning accuracy can be achieved with only $10$ features. For Fig. \ref{fig:compare_dnn}, the corresponding loss tendencies regarding training epochs are shown in Appendix \ref{appendix3}.

\subsection{Energy Efficiency vs. Learning Accuracy}

\begin{figure}[t] 
\centering
\centering
\subfigure[SVM classifier]{\includegraphics[width=8cm]{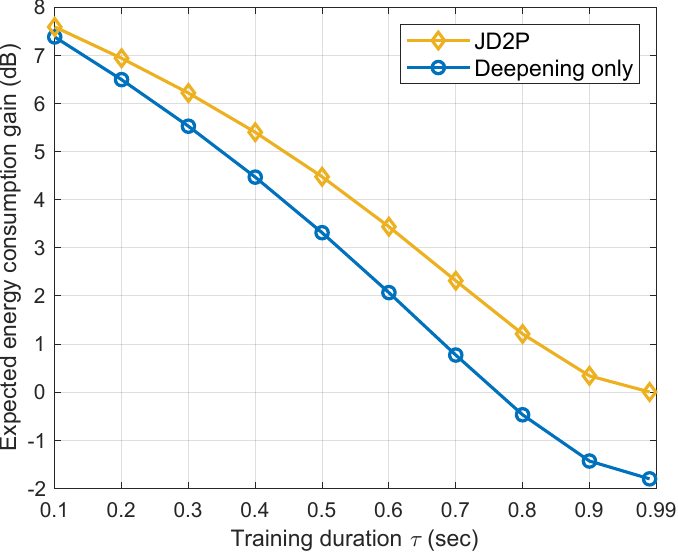}}
\subfigure[DNN classifier]{\includegraphics[width=8cm]{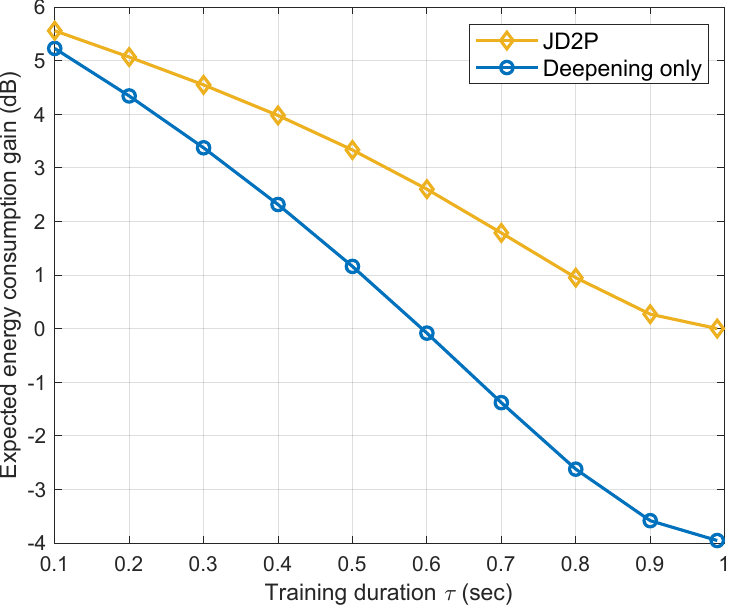}}
\vspace{-3pt}
\caption{Effect of the prefetching duration on the expected energy consumption gain that is normalized by the OSC. Except for the training duration being extremely large, the gain in energy consumption has a positive value. We consider (a) the SVM classifier ($p_{\mathrm{th}} \in [0.95, 0.995]$) and (b) the DNN classifier ($z_{\mathrm{th}} \in [0.01,0.09]$). }
\label{fig:energy_gain_tau}
\end{figure} 

\begin{figure}[t] 
\centering
\centering
\subfigure[SVM classifier]{\includegraphics[width=8.1cm]{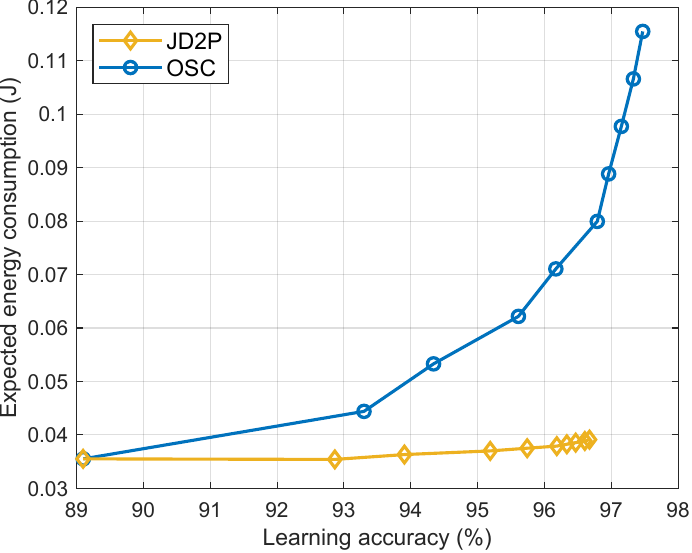}}
\subfigure[DNN classifier]{\includegraphics[width=8cm]{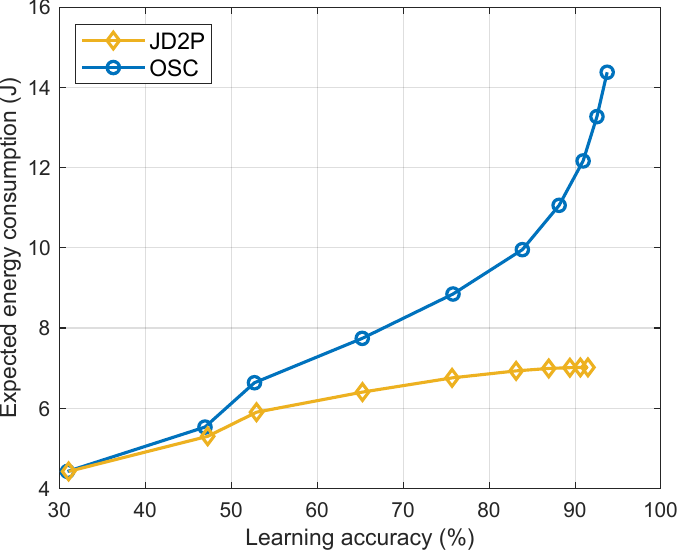}}
\caption{The effect of the number of total rounds, $N \in [1,10]$, on the expected energy consumption gain. The parameters are set as $p_{\mathrm{th}} = 0.98$, $z_{\mathrm{th}} = 0.03$ and $\tau = 0.5$s. }
\label{fig:energy_gain_N}
\vspace{-10pt}
\end{figure} 

In this subsection, we investigate the tradeoff between energy consumption and learning accuracy and analyze the effect of various relevant parameters.  
First, the expected energy consumption (in Joule) versus the error rate (in \%) is plotted in Fig. \ref{fig:energy_tradeoff}. It is shown that the proposed JD2P always consumes less energy than that of OSC with $10$ features, namely, $4.1$dB and $2.5$dB energy gains for SVM and DNN classifiers, respectively. When the error-tolerance parameters $p_{\mathrm{th}}$ and $z_{\mathrm{th}}$ are beyond $0.99$ and $0.07$, respectively, the learning accuracy becomes better than that of OSC, as explained before (see Fig. \ref{fig:tradeoff} and Section. \ref{subsec:tradeoff-deepening}). 
Additionally, the expected energy consumption of deepening only can occasionally exceed that of OSC by up to $1.6$ Joule (see Fig. \ref{fig:energy_tradeoff_dnn}), which occurs when the available transmit time is insufficient for the deepening~process.


Besides, it is shown that the energy consumption is significantly reduced with J2DP compared to the case using data deepening only without prefetching. The effectiveness of data prefetching is extensively demonstrated in Fig. \ref{fig:energy_gain_tau}, plotting the curves of the energy consumption gain against the prefetching duration of $\tau$. Here, the gain is defined as the expected energy consumption normalized by OSC. First, the energy gain of JD2P is always higher than that of data deepening only by a sophisticated control of prefetching data in Section. \ref{section:optimal_data_prefetching}. On the other hand, when compared with  OSC, the energy gain of JD2P decreases as $\tau$ increases. In other words, a larger training duration compels more data samples to be prefetched, wasting more energy since many prefetched data samples are likely to become CCSs while not being used for the following~training.

\begin{figure}[t] 
\centering
\centering
\subfigure[SVM classifier]{\includegraphics[width=8.3cm]{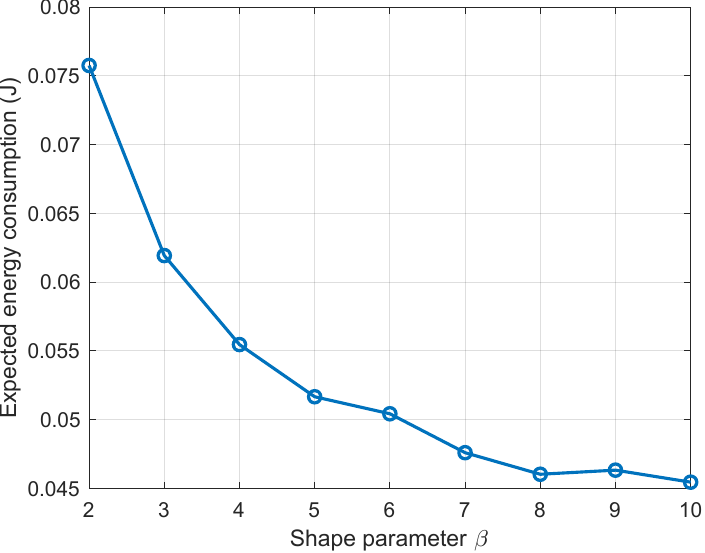}\label{fig:energy_tradeoff_svm}}
\subfigure[DNN classifier]{\includegraphics[width=8cm]{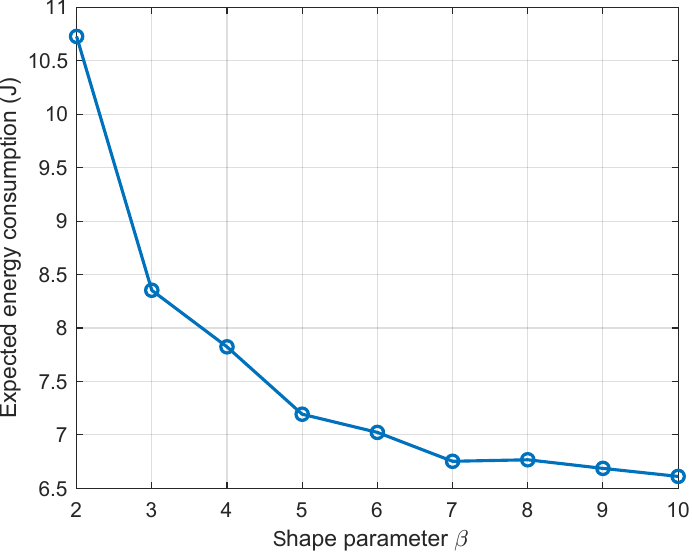}\label{fig:energy_tradeoff_dnn}}
\caption{The expected energy consumption decreases as the channel randomness decreases. Each point represents the average value over error-tolerance parameters with $p_{\mathrm{th}} \in [0.95, 0.995]$ and $z_{\mathrm{th}} \in [0.01,0.09]$. The training duration is set as $\tau = 0.5$s.}
\label{fig:shape_parameter}
\vspace{-10pt}
\end{figure} 

Fig. \ref{fig:energy_gain_N} shows the energy consumptions of J2DP and OSC when the number of rounds increases. While the energy consumption of OSC exponentially increases, that of J2DP tends to be stationary for both binary SVM and multi-class DNN classifiers. With data deepening, JD2P gradually reduces the number of features offloaded in each round, and its expected energy consumption converges to a specific value even though the number of rounds increases. It implies that JD2P automatically decides whether offloading is terminated through the classifier trained so far.

\subsection{Effect of the Channel Distribution} \label{subsec:effec_of_channel}

{This subsection investigates the relation between the randomness of the channel and the expected energy consumption. The shape parameter $\beta$ in the gamma distribution in Sec. \ref{subsubsec:setting_offloading} controls the randomness of the channel such that a higher $\beta$ makes the channel less fluctuate. For a fair comparison, we fix the channel’s mean to $1$. As shown in Fig. \ref{fig:shape_parameter}, the expected energy consumption decreases when $\beta$ decreases, which empirically confirms our discussion in Remark 6.}

\section{Concluding Remarks}\label{Section:concluding remarks}
This study has explored the problem of a multi-round offloading technique for energy-efficient edge learning. Two criteria for achieving energy efficiency are 1) to reduce the amount of offloaded data and 2) to extend the offloading duration. The proposed JD2P was designed in the sense of addressing both criteria by integrating data deepening and prefetching techniques. The former measures feature-by-feature data importance, while the latter optimizes the amount of prefetched data to avoid wasting energy. The applicability of JD2P has been demonstrated for a binary classifier based on SVM and a multi-class classifier based on DNN. Through comprehensive analytic and numerical studies, JD2P has shown a significant reduction in the expected energy consumption compared to several benchmarks. 

This work can be extended to several challenging research directions. 
First, the current J2DP based on supervised learning can be re-designed for self-supervised or semi-supervised learning approaches.    
Second, the feature importance can also be applied to design a scheduling algorithm in federated edge learning,  prioritizing {IoT} devices based on the relevance and informativeness of their feature representations. Last, it is interesting to integrate J2DP with state-of-art network techniques such as network slicing and virtualization.



\appendix
\subsection{Proof of Proposition \ref{proposition 1}}\label{appendix1}
Define the Lagrangian function for \ref{Problem:upper bound} as
\begin{align}
  L = \frac{p_k^\ell}{h_{k}\tau_k^{\ell-1}} + \frac{\nu}{t_{k+1}^{\ell-1}} \left(\left(s_k-p_k\right)\rho_k + \frac{\ell}{2}\right)^\ell + \eta (p_k - s_k),\nonumber
\end{align}
where $\eta$ is a Lagrangian multiplier. Since \ref{Problem:upper bound} is a convex optimization, the following KKT conditions are necessary and sufficient for optimality:
\begin{subequations}\label{eq:kkt}
    \begin{gather}
        \frac{\ell p_k^{\ell-1}}{h_{k}\tau_k^{\ell-1}} - \frac{\ell \nu \rho_k }{t_{k+1}^{\ell-1}} \left(\left(s_k-p_k\right)\rho_k + \frac{\ell}{2}\right)^{\ell-1} +\eta\geq 0, \label{kkt1} \\ 
        p_k \left(\frac{\ell p_k^{\ell-1}}{h_{k}\tau_k^{\ell-1}} - \frac{\ell \nu \rho_k }{t_{k+1}^{\ell-1}} \left(\left(s_k-p_k\right)\rho_k + \frac{\ell}{2}\right)^{\ell-1} +\eta\right) = 0, \label{kkt2} \\ 
        \eta \left(p_k-s_k\right) = 0. \label{kkt3}
    \end{gather}
\end{subequations}
First, if $\eta$ is positive, then $p_k$ should be equal to $s_k$ due to the slackness condition of \eqref{kkt3}, making the left-hand side (LHS) of \eqref{kkt2} strictly positive. In other words, the optimal multiplier $\eta$ is zero to satisfy \eqref{kkt2}. Second, with $p_k=0$, the LHS of condition \eqref{kkt1} is always strictly negative unless $\rho_k=0$. As a result, given $\rho_k>0$, $p_k$ should be strictly positive and satisfy the following equality condition: 
\begin{align} \label{eq:last_eq}
  \frac{\ell p_k^{\ell-1}}{g_{k}\tau_k^{\ell-1}} - \frac{\ell \nu \rho_k }{t_{k+1}^{\ell-1}} \left(\left(s_k-p_k\right)\rho_k + \frac{\ell}{2}\right)^{\ell-1}=0. 
\end{align}
Solving \eqref{eq:last_eq} leads to the optimal solution of \ref{Problem:upper bound}, which completes the proof of this proposition.

\begin{figure}[t] 
\centering
\centering
\subfigure[SVM classifier]{\includegraphics[width=8.1cm]{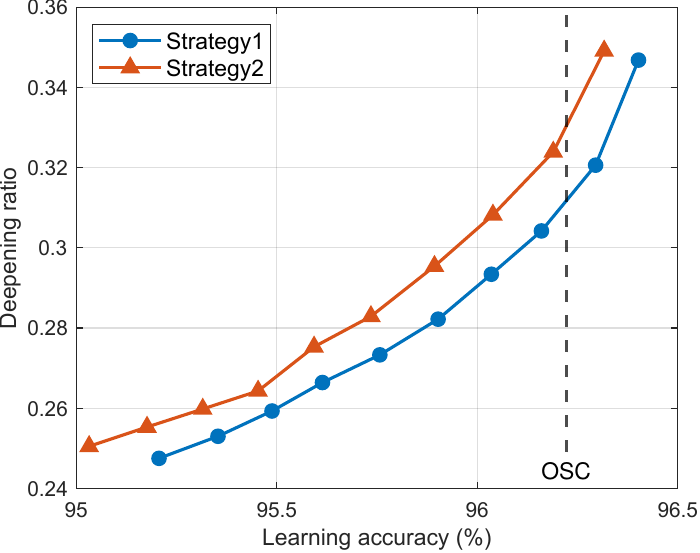}\label{fig:tradeoff_svm}}
\subfigure[DNN classifier]{\includegraphics[width=8cm]{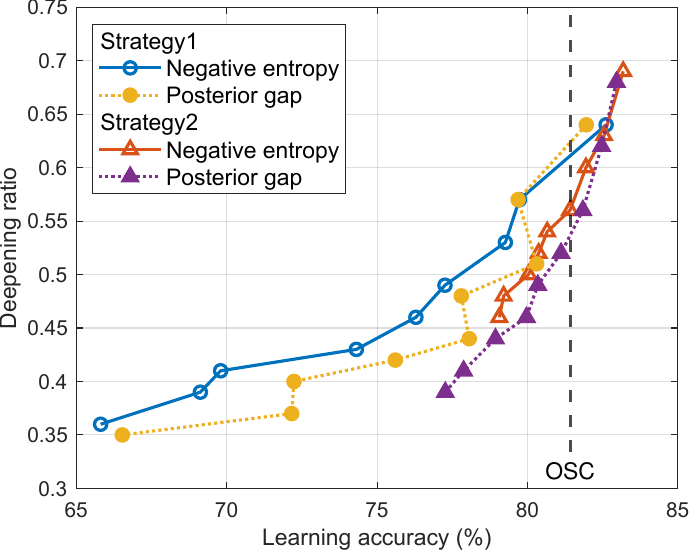}\label{fig:tradeoff_dnn}}
\caption{Data deepening exhibits a tradeoff between the learning accuracy and the deepening ratio. The performance is evaluated using the Fashion MNIST dataset. The deepening ratio is obtained by comparing the amount of data samples with the OSC. We consider (a) the SVM classifier ($p_{\mathrm{th}} \in [0.95, 0.995]$) and (b) the DNN classifier ($z_{\mathrm{th}} \in [0.01,0.09]$).}
\label{fig:fashion_mnist_tradeoff}
\vspace{-15pt}
\end{figure}

\subsection{Extension to Fashion MNIST}\label{appendix2}

{We have verified data deepening using the Fashion MNIST dataset, which has the same data size and format as the MNIST dataset. The entire training set includes $7 \cdot 10^4$ images, which are divided into $6 \cdot 10^4$ and $10^4$ images for training and testing, respectively. The experiment is performed in the same manner, specified in Sec. \ref{Section:Simulation}. As shown in Fig. \ref{fig:fashion_mnist_tradeoff}, the tradeoff between the deepening ratio and learning accuracy tends to be similar to MNIST. One interesting observation is that strategy 2 performs worse than strategy 1 when training the SVM classifier, which is different from the case with the MNIST dataset. The reason is that the Fashion MNIST dataset is more complex, making it difficult for a linear SVM to train all data samples of different data depths at once. However, the DNN can be trained well using data samples with varying data depths, resulting in higher performance in strategy 2. }

\subsection{Learning Loss vs. Epoch} \label{appendix3}

\begin{figure}[t] 
\centering
\centering
\subfigure[$z_\mathrm{th}=0.01$]{\includegraphics[width=8cm]{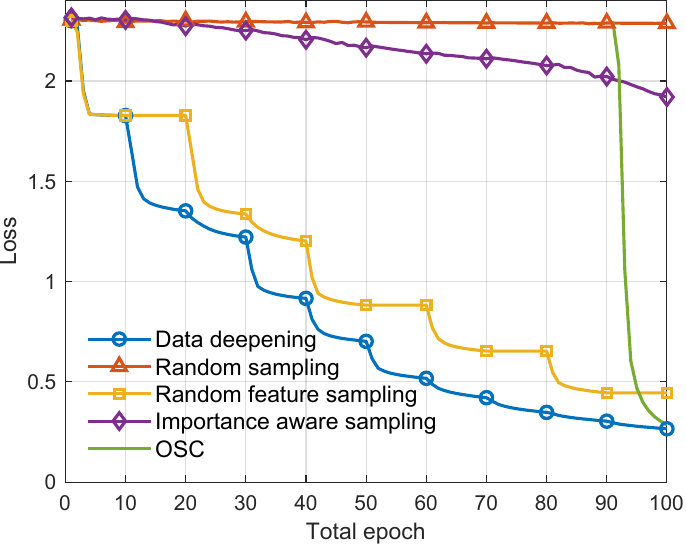}}
\subfigure[$z_\mathrm{th}=0.09$]{\includegraphics[width=8cm]{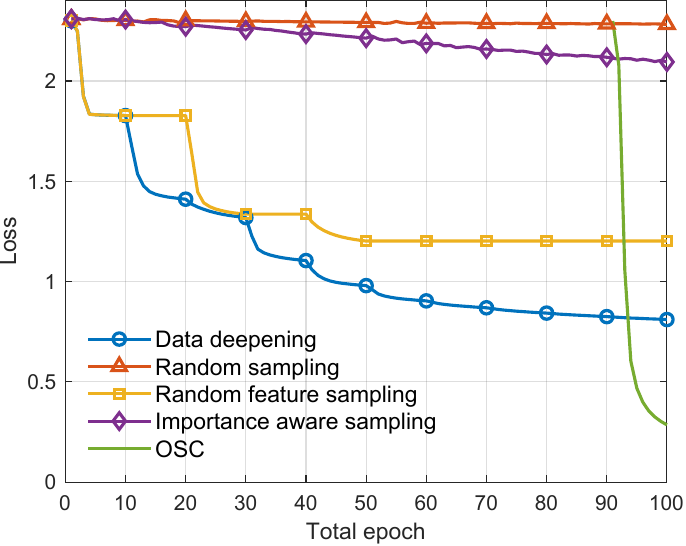}}
\vspace{-5pt}
\caption{The loss of data deepening is compared with several benchmarks at every epoch. Benchmarks use the same number of data samples that data deepening uses, except OSC. We consider error tolerance parameter (a) $z_\mathrm{th} = 0.01$ and (b) $z_\mathrm{th} = 0.09$.}
\label{fig:epoch_loss}
\vspace{-10pt}
\end{figure} 

Fig. 14 shows the loss tendency according to epochs, where ten epochs correspond to one round for data deepening. We compare the proposed data deepening with several benchmarks specified in Sec. VII.A. Note that the loss of OSC is presented only in the range of 90 - 100 epochs, representing the final round since OSC trains the classifier at once after offloading all data samples. On the other hand, the number of data samples that the benchmarks can offload is equivalent to data deepening for a fair comparison. One observes that the other benchmarks' losses are higher than those of data deepening since only data deepening considers the importance of the feature, achieving a lower training loss with faster convergence.

\bibliographystyle{ieeetran}
\bibliography{refer}

\section{Biography}

\begin{IEEEbiography}[{\includegraphics[width=1in,height=1.25in,clip,keepaspectratio]{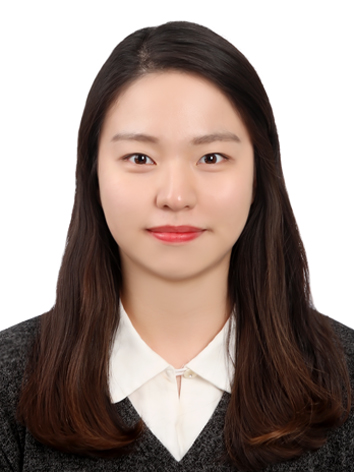}}]{Sujin Kook}
received the B.S. degree in electrical and electronic engineering from Chung-Ang University, Seoul, Republic of Korea, in 2020. She is currently pursuing the Ph.D. degree in electrical and electronic engineering with Yonsei University, Seoul, South Korea. Her research interests include importance-based edge learning for wireless communications, remote robot control, network slicing, private 5G and Open RAN, and related implementation.
\end{IEEEbiography}

\begin{IEEEbiography}[{\includegraphics[width=1in,height=1.25in,clip,keepaspectratio]{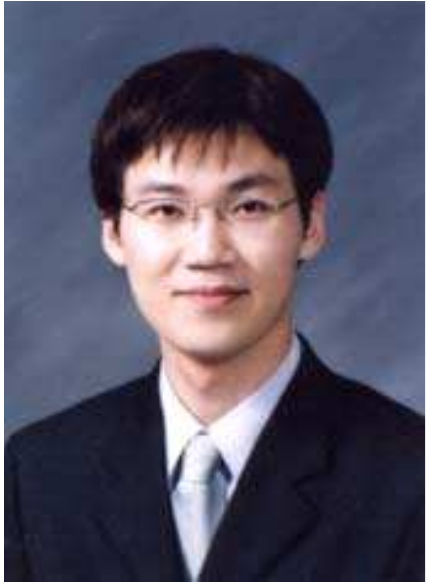}}]{Won-Yong Shin}
received the B.S. degree in electrical engineering from Yonsei University, Seoul, Republic of Korea, in 2002. He received the M.S. and the Ph.D. degrees in electrical engineering and computer science from the Korea Advanced Institute of Science and Technology (KAIST), Daejeon, Republic of Korea, in 2004 and 2008, respectively. In May 2009, Dr. Shin joined the School of Engineering and Applied Sciences, Harvard University, Cambridge, MA USA, as a Postdoctoral Fellow and was promoted to a Research Associate in October 2011. From March 2012 to February 2019, he was a Faculty Member (with {\it tenure}) of the Department of Computer Science and Engineering, Dankook University, Yongin, Republic of Korea. Since March 2019, he has been with the School of Mathematics and Computing (Computational Science and Engineering), Yonsei University, Seoul, Republic of Korea, where he is currently a Full Professor. He has also been with the Graduate School of Artificial Intelligence at Pohang University of Science and Technology (POSTECH) as an Adjunct Professor since September 2022. His research interests are in the areas of information theory, mobile computing, data mining, and machine learning.
From 2014 to 2018, Dr. Shin served as an Associate Editor of the {\it IEICE Transactions on Fundamentals of Electronics, Communications and Computer Sciences}. Since 2022, he has served as an Associate Editor of the {\it ICT Express}. He also served as an Organizing Committee member for the {\it 2023/2024 IEEE Consumer Communications \& Networking Conference} and the {\it 2015 IEEE Information Theory Workshop}. Additionally, he served as a Technical Committee member or a Chair for various conferences. He received the Bronze Prize of the Samsung Humantech Paper Contest (2008/2022), the KICS Haedong Young Scholar Award (2016), and the ICT Express Best Guest Editor Award (2021).
\end{IEEEbiography}

\begin{IEEEbiography}[{\includegraphics[width=1in,height=1.25in,clip,keepaspectratio]{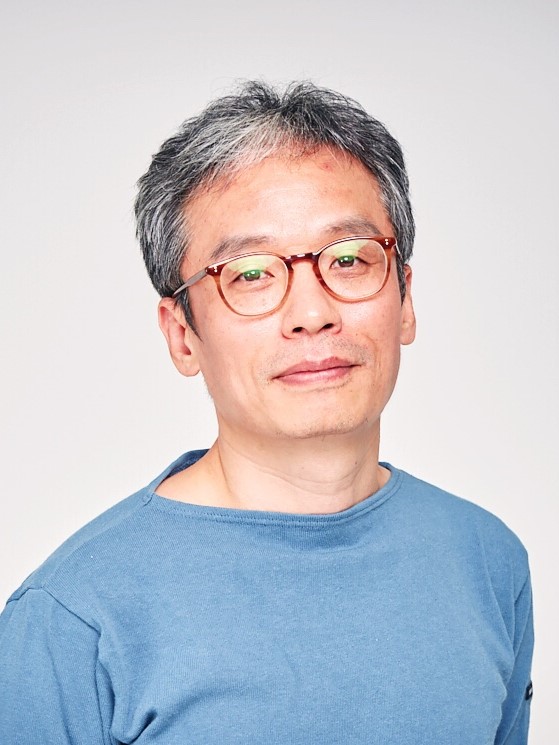}}]{Seong-Lyun Kim}
is a Professor at the School of EEE, Yonsei University, Seoul, Korea. He recently led the Smart Factory Committee of 5G Forum, Korea. He was an Assistant/Associate Professor of Radio Communication Systems, Royal Institute of Technology (KTH), Stockholm, Sweden. He had visiting positions in Helsinki University of Technology (now Aalto), Finland, the KTH Center for Wireless Systems, and the Graduate School of Informatics, Kyoto University, Japan. He served as an editorial board member of IEEE Transactions on Vehicular Technology, IEEE Communications Letters, Elsevier Control Engineering Practice, Elsevier ICT Express, and Journal of Communications and Network. He served as a guest editor of IEEE Wireless Communications and IEEE Network for wireless communications in networked robotics, and IEEE Journal on Selected Areas in Communications. His research interests include radio resource management, AI/ML in wireless networks, collective intelligence, and robotic networks. He is a co-recipient of IEEE VTC best paper award, IEEE Dyspan best demo award, and more recently, IEEE Heinrich Hertz Award.
\end{IEEEbiography}


\begin{IEEEbiography}[{\includegraphics[width=1in,height=1.25in,clip,keepaspectratio]{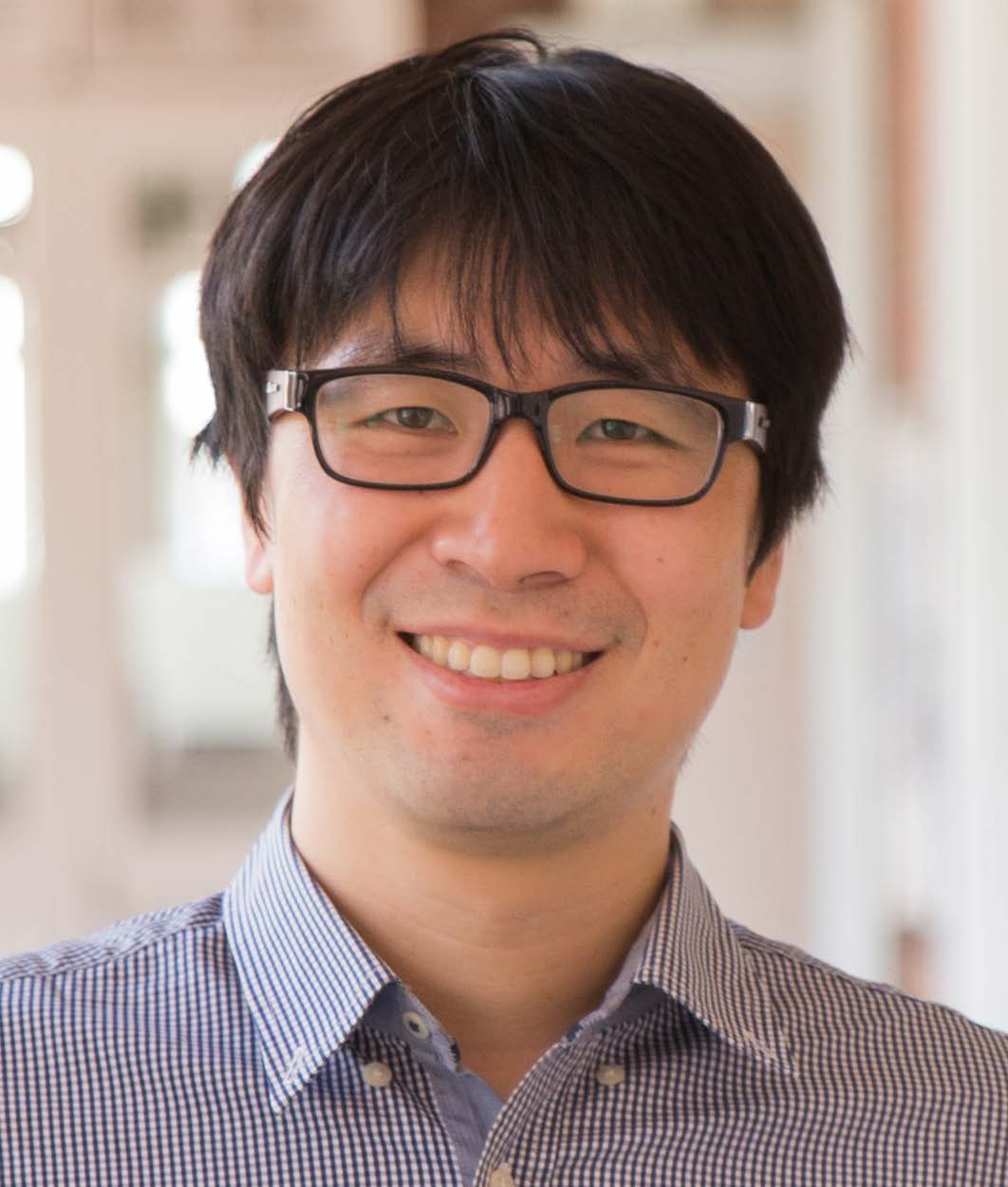}}]{Seung-Woo Ko}
(Senior Member, IEEE) received the B.S., M.S., and Ph.D. degrees from the School of Electrical and Electronic Engineering, Yonsei University, South Korea, in 2006, 2007, and 2013, respectively. Since 2022, he has been an Associate Professor with the Department of Smart Mobility Engineering at Inha University. Before joining Inha University, he was an Assistant Professor at Korea Maritime and Ocean University (KMOU), Busan, South Korea, from 2019 to 2021, a Senior Researcher with LG Electronics, South Korea, from 2013 to 2014, and a Post-Doctoral Researcher with Yonsei University, South Korea in 2015, and The University of Hong Kong (HKU) from 2016 to 2019. His research interests include intelligent wireless communications and networking for 6G, with particular emphasis on semantic communications, integrated sensing and communications, and radio-based positioning.
\end{IEEEbiography}

\end{document}